\documentclass[10pt,twocolumn,letterpaper]{article}

\usepackage{cvpr}
\usepackage{times}
\usepackage{epsfig}
\usepackage{graphicx}
\usepackage{amsmath}
\usepackage{amssymb}
\usepackage{multirow}
\usepackage{booktabs}
\usepackage[table]{xcolor}
\usepackage{threeparttable}
\usepackage{colortbl}
\usepackage{algorithm}
\usepackage{algorithmic}
\usepackage{amsmath,bm}
\usepackage{multirow}
\usepackage{booktabs}
\usepackage[table]{xcolor}
\usepackage{amsmath}
\usepackage{amssymb}
\usepackage{color}
\usepackage{times}
\usepackage{epsfig}
\usepackage{booktabs,multirow}
\usepackage{graphics}
\usepackage{threeparttable}
\usepackage{color}
\usepackage[normalem]{ulem}
\usepackage{multirow}
\usepackage{float}
\usepackage{amsfonts}
\usepackage{bm}
\usepackage{subfig}
\usepackage{enumitem}
\usepackage[numbers]{natbib}
\usepackage{array}
\usepackage[table]{xcolor}
\usepackage{colortbl}
\definecolor{bblue}{rgb}{0,150,230}
\definecolor{mygray}{gray}{.9}
\usepackage{bm}

\newcolumntype{I}{!{\vrule width 1pt}}

\definecolor{ggray}{RGB}{127,127,127}

\makeatletter
\newcommand{\thickhline}{%
    \noalign {\ifnum 0=`}\fi \hrule height 1pt
    \futurelet \reserved@a \@xhline
}
\makeatother
\newcommand{\tabincell}[2]{\begin{tabular}{@{}#1@{}}#2\end{tabular}}
\usepackage[pagebackref=true,breaklinks=true,letterpaper=true,colorlinks,bookmarks=false]{hyperref}

\cvprfinalcopy % *** Uncomment this line for the final submission

\def\cvprPaperID{1092} % *** Enter the CVPR Paper ID here

\ifcvprfinal\fi
\begin{document}

%%%%%%%%% TITLE
\title{Hierarchical Human Parsing with Typed Part-Relation Reasoning}

\author{Wenguan Wang$^{1,2}$\thanks{The first two authors contribute equally to this work.}~,\hspace{1pt} Hailong Zhu$^{3*}$~, \hspace{1pt} Jifeng Dai$^{4}$~,\hspace{1pt} Yanwei Pang$~^{3}$\thanks{Corresponding author: \textit{Yanwei Pang}. }~, \hspace{1pt} Jianbing Shen$^{2}$~, \hspace{1pt}  Ling Shao$~^{2}$~\\
	\small{$^1$} \small ETH Zurich, Switzerland \hspace{0pt} \hspace{2pt} \small{$^2$} \small Inception Institute of Artificial Intelligence, UAE \\
	\!\!\!\!\!\!\!\!\small{$^3$} \small Tianjin Key Laboratory of Brain-inspired Intelligence Technology, School of Electrical and Information Engineering, Tianjin University, China	\\\small{$^4$} \small SenseTime Research  \hspace{3pt} \\
	{\tt\small \{wenguanwang.ai,hlzhu2009\}@gmail.com}\\
	{\tt\small \url{https://github.com/hlzhu09/Hierarchical-Human-Parsing}}
\vspace{-15pt}
}

\maketitle
%\thispagestyle{empty}

%%%%%%%%% ABSTRACT
\begin{abstract}
\vspace{-5pt}
Human parsing is for pixel-wise human semantic understanding. As human bodies are underlying hierarchically structured, how to model human structures is the central theme in this task. Focusing on this, we seek to simultaneously exploit the representational capacity of deep graph networks and the hierarchical human structures. In particular, we provide following two contributions. First,
three kinds of part relations, \ie, decomposition, composition, and dependency, are, for the first time, completely and precisely described by three distinct relation networks. This is in stark contrast to previous parsers, which only focus on a portion of the relations and adopt a type-agnostic relation modeling strategy. More expressive relation information can be captured by explicitly imposing the parameters in the relation networks to satisfy the specific characteristics of different relations. %, thus further benefiting the parsing results.
Second, previous parsers largely ignore the need for an approximation algorithm over the loopy human hierarchy, while we instead address an iterative reasoning process, by assimilating generic message-passing networks with their edge-typed, convolutional counterparts. With these efforts, our parser lays the foundation for more sophisticated and flexible human relation patterns of reasoning. Comprehensive experiments on five datasets demonstrate that our parser sets a new state-of-the-art on each.

%Integrating human part relations into human parsers are extremely useful for human semantic parsing.  However, part relations, \ie, composition, decomposition, and dependency, reside on human bodies yield distinct relation rules, Explicitly and precisely capture the compositional/decompositional relations between constituent and entity parts and help with the dependency learning over the kenaidcally connected parts.
%to obtain a hierarchical human body representation with increasing levels of human structure modeling
%This paper seeks to simultaneously exploit
%the representational capacity of deep networks and the compositional linguistic structure of questions
%We evaluate our approach on two challenging datasets for visual question answering, achieving state-of-the-art results undergo distinguish  very nature  very nature

%The resulting architecture
% Just as biology uses nature and nurture cooperatively, we reject the false choice between “hand-engineering” and “end-to-end” learning, and instead advocate for an approach which benefits from their complementary strengths
\end{abstract}

%%%%%%%%% BODY TEXT
\vspace{-8pt}
\section{Introduction}
\vspace{-2pt}
Human parsing involves segmenting human bodies into semantic parts, \eg, head, arm, leg, \etc. It has attracted tremendous attention in the literature, as it enables fine-grained human understanding and finds a wide spectrum of human-centric applications, such as human behavior analysis~\cite{qi2018learning,fan2019understanding}, %clothing recognition,
human-robot interaction~\cite{fang2019graspnet}, and many others.

%Intelligent applications benefit from structured
%knowledge about the entities and relations in
%their domains. For example While these knowledge bases are often carefully
%curated, they are far from complete. In non-static
%domains, new facts become true or are discovered
%at a fast pace, making the manual expansion of
%knowledge bases impractical. Extracting relations
%from a text corpus (Mintz et al., 2009; Surdeanu
%et al., 2012; Poon et al., 2015) or inferring facts
%from the relationships among known entities (Lao
%and Cohen, 2010) are thus important approaches
%for populating existing knowledge bases
%%%%%%%%%%%%%%%%%%% Figure 1 %%%%%%%%%%%%%%%%%%%%%%
\begin{figure}[t]
%%tr = 0.006, ts = 0.008
  \centering
      \includegraphics[width=0.99\linewidth]{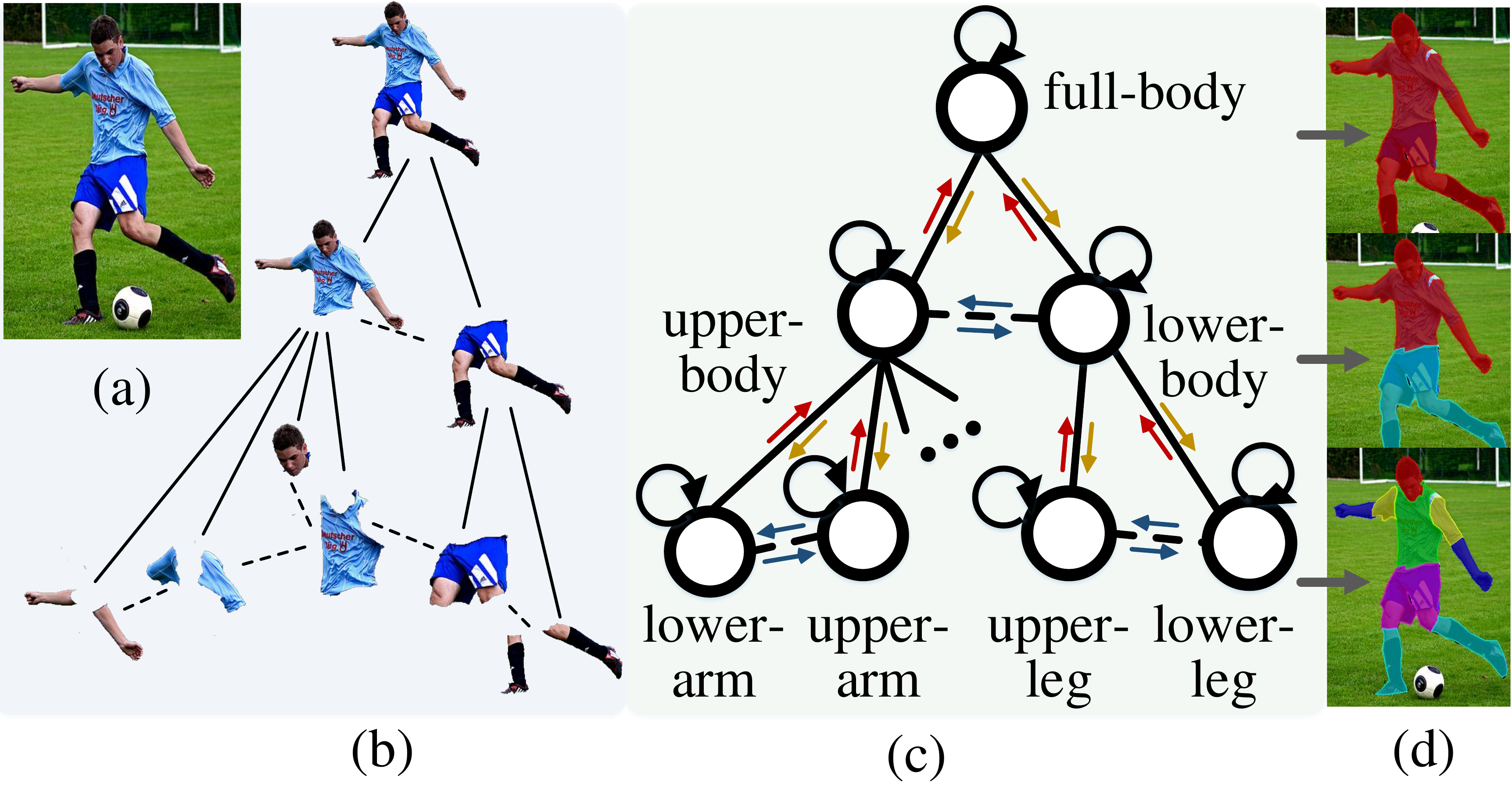}
\vspace{-10pt}
\captionsetup{font=small}
\caption{\small \!\textbf{Illustration of our hierarchical human parser.} (a) Input image. (b) The human hierarchy in (a), where \protect\includegraphics[scale=0.16]{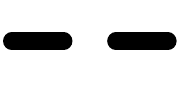}$_{\!}$ indicates dependency relations and \protect\includegraphics[scale=0.16]{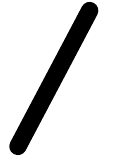} is de-/compositional relations. (c)~In our parser, three distinct relation networks are designed for addressing the specific characteristics of different part relations, \ie, \protect\includegraphics[scale=0.16]{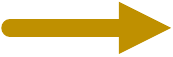}, \protect\includegraphics[scale=0.16]{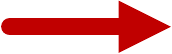}, and \protect\includegraphics[scale=0.16]{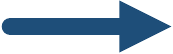} stand for decompositional, compositional, and dependency  relation networks, respectively. Iterative inference (\protect\includegraphics[scale=0.16]{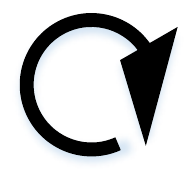}) is performed for better approximation. For visual clarity, some nodes are omitted. (d) Our hierarchical parsing results.
}
\label{fig:overview}
\vspace{-16pt}
\end{figure}

Human bodies present a highly structured hierarchy and body parts inherently interact with each other. As shown in Fig.\!~\ref{fig:overview}(b), there are different relations between parts~\cite{4587787,wang2018attentive,7990516}: \textbf{decompositional} and \textbf{compositional} relations (full line:\includegraphics[scale=0.16]{figs/de-com.pdf}) between constituent and entire parts (\eg, \{\textit{upper body}, \textit{lower body}\} and \textit{full body}), and \textbf{dependency} relations (dashed line:\includegraphics[scale=0.16]{figs/depen.pdf}) between kinematically connected parts (\eg, \textit{hand} and \textit{arm}). Thus the central problem in human parsing is how to model such relations. Recently, numerous structured human parsers have been proposed~\cite{xia2016pose,fang2018weakly,gong2017look,xia2017joint,nie2018mutual,Zhu2018ProgressiveCH,wang2019CNIF,Gong_2019_CVPR}. Their notable successes indeed demonstrate the benefit of exploiting the structure in this problem. However, three major limitations in human structure modeling are still observed. \textbf{(1)} The structural information utilized is typically weak and relation types studied are incomplete. Most efforts~\cite{xia2016pose,fang2018weakly,gong2017look,xia2017joint,nie2018mutual} directly encode human pose information into the parsing model, causing them to suffer from trivial structural information, not to mention the need of extra pose annotations. In addition, previous structured parsers focus on only one or two of the aforementioned part relations, not all of them. For example, \cite{Gong_2019_CVPR} only considers dependency relations, and \cite{Zhu2018ProgressiveCH} relies on decompositional relations. \textbf{(2)} Only a single relation model is learnt to reason different kinds of relations, without considering their essential and distinct geometric constraints. Such a relation modeling strategy is over-general and simple; do not seem to characterize well the diverse part relations. \textbf{(3)} According to graph theory, as the human body yields a complex, cyclic topology, an iterative inference is desirable for optimal result approximation. However, current arts~\cite{gong2017look,xia2017joint,nie2018mutual,Zhu2018ProgressiveCH,wang2019CNIF} are primarily built upon an immediate, feed-forward prediction scheme.% without iterative approximate inference.

To respond to the above challenges and enable a deeper understanding of human structures, we develop a unified, structured human parser that precisely describes a more complete set of part relations, and efficiently reasons structures with the prism of a message-passing, feed-back inference scheme. To address the first two issues, we start with an in-depth and comprehensive analysis on three essential relations, namely decomposition, composition, and dependency. Three distinct relation networks (\includegraphics[scale=0.16]{figs/R-decom.pdf}, \includegraphics[scale=0.16]{figs/R-com.pdf}, and \includegraphics[scale=0.16]{figs/R-depen.pdf} in Fig.\!~\ref{fig:overview}(c)) are elaborately designed and imposed to explicitly satisfy the specific, intrinsic relation constraints. Then, we construct our parser as a tree-like, end-to-end trainable graph model, where the nodes represent the human parts, and edges are built upon the relation networks. For the third issue, a modified, relation-typed convolutional message passing procedure (\includegraphics[scale=0.16]{figs/itera.pdf} in Fig.\!~\ref{fig:overview}(c)) is performed over the human hierarchy, enabling our method to obtain better parsing results from a global view.  All components, \ie, the part nodes, edge (relation) functions, and message passing modules, are fully differentiable, enabling our whole framework to be end-to-end trainable and, in turn, facilitating learning about parts, relations, and inference algorithms.

More crucially, our structured human parser can be viewed as an essential variant of message passing neural networks (MPNNs)~\cite{gilmer2017neural,velickovic2017graph}, yet significantly differentiated
in two aspects. (1) Most previous MPNNs are edge-type-agnostic, while ours addresses relation-typed structure reasoning with a higher expressive capability. (2) By replacing the Multilayer Perceptron (MLP) based MPNN units with convolutional counterparts, our parser gains a spatial information preserving property, which is desirable for such a pixel-wise prediction task.

We extensively evaluate our approach on five standard human parsing datasets~\cite{gong2017look,xia2017joint,luo2013pedestrian,liang2015deep,Luo_2018_TGPnet}, achieving state-of-the-art performance on all of them (\S\ref{sec:qresults1}). In addition, with ablation studies for each essential component in our parser (\S\ref{sec:ablation}), three key insights are found: (1) Exploring different relations reside on human bodies is valuable for human parsing. (2) Distinctly and explicitly modeling different types of relations can better support human structure reasoning. (3) Message passing based feed-back inference is able to reinforce parsing results. %We believe our work is an important step towards making well-designed MPNNs  Symbolic format does make the tasks easier but still nontrivial.
\vspace{-3pt}
\section{Related Work}
\vspace{-3pt}
\noindent\textbf{Human parsing:}  Over the past decade, active
research has been devoted towards pixel-level human semantic understanding. Early approaches tended to leverage
image regions~\cite{liu2014fashion,yamaguchi2012parsing,yang2014clothing}, hand-crafted features~\cite{wang2011blocks,chen2014detect}, part templates~\cite{bo2011shape,dong2013deformable,dong2014towards} and human keypoints~\cite{yamaguchi2013paper,liu2014fashion,yamaguchi2012parsing,yang2014clothing}, and typically explored certain heuristics over human body configurations~\cite{chen2006composite,dong2013deformable,dong2014towards} in a CRF~\cite{yamaguchi2013paper,ladicky2013human}, structured model~\cite{yamaguchi2012parsing,dong2013deformable}, grammar model~\cite{chen2006composite,4587787,dong2014towards}, or generative model~\cite{eslami2012generative,rauschert2012generative} framework. %Due to the lack of a robust learning strategy and weak representability of hard-crafted features, their performance is limited in realistic scenes.
Recent advance has been driven by the streamlined designs of deep learning architectures. Some pioneering efforts revisit classic template matching strategy~\cite{liang2015deep,liu2015matching}, address local and global cues~\cite{liang2015human}, or use tree-LSTMs to gather structure information~\cite{liang2016semantic,liang2016semantic2}. However, due to the use of superpixel~\cite{liang2015human,liang2016semantic,liang2016semantic2} or HOG feature~\cite{luo2013pedestrian}, they are fragmentary and time-consuming. Consequent attempts thus follow a more elegant FCN architecture, addressing  multi-level cues~\cite{chen2016attention,xia2016zoom}, feature aggregation~\cite{Luo_2018_TGPnet,zhao2017self,liu2017surveillance},  adversarial learning~\cite{zhao2018understanding,luo2018macro,liu2018cross}, or cross-domain knowledge~\cite{liu2018cross,xu2018srda,Gong_2019_CVPR}. To further explore inherent structures, numerous approaches~\cite{xia2016pose,zhao2017self,gong2017look,xia2017joint,fang2018weakly,nie2018mutual} choose to straightforward encode pose information into the parsers, however, relying on off-the-shelf pose estimators~\cite{fang2018learning,fang2017rmpe} or additional annotations. Some others consider top-down~\cite{Zhu2018ProgressiveCH} or multi-source semantic~\cite{wang2019CNIF} information over hierarchical human layouts. Though impressive, they ignore iterative inference and seldom address explicit relation modeling, easily suffering from weak expressive ability and risk of sub-optimal results.
% produced significant results

With the general success of these works, we make a further step towards more precisely describing the different relations residing on human bodies, \ie, decomposition, composition, and dependency,  and addressing iterative, spatial-information preserving inference over human hierarchy.

% As a part of the huge graph learning family,
\noindent\textbf{Graph neural networks (GNNs):}GNNs have a rich history (dating back to~\cite{scarselli2008graph}) and became a veritable explosion in research community over the last few years~\cite{hamilton2017representation}. GNNs effectively learn graph representations in an end-to-end manner, and can generally be divided into two broad classes: Graph Convolutional Networks (GCNs) and Message Passing Graph Networks (MPGNs). The former~\cite{duvenaud2015convolutional,niepert2016learning,kipf2016semi} directly extend classical CNNs to %structured,
non-Euclidean data. Their simple architecture promotes their popularity, %massive graphs s (e.g., millions of nodes)
while limits their modeling capability for complex structures~\cite{hamilton2017representation}. MPGNs~\cite{gilmer2017neural,zheng2019reasoning,velickovic2017graph,wang2019zero} parameterize all the nodes, edges, and information fusion steps in graph learning, leading to more complicated yet flexible architectures.% We refer the reader to XX for an exhaustive literature review on the topic.

Our structured human parser, which falls in the second category, can be viewed as an early attempt to explore GNNs in the  area of human parsing. In contrast to conventional MPGNs, which are mainly MLP-based and edge-type-agnostic, we provide a spatial information preserving and relation-type aware graph learning scheme.

%%%%%%%%%%%%%%%%%%% Figure 2%%%%%%%%%%%%%%%%%%%%%%
\begin{figure*}[t]
%%tr = 0.006, ts = 0.008
  \centering
      \includegraphics[width=1 \linewidth]{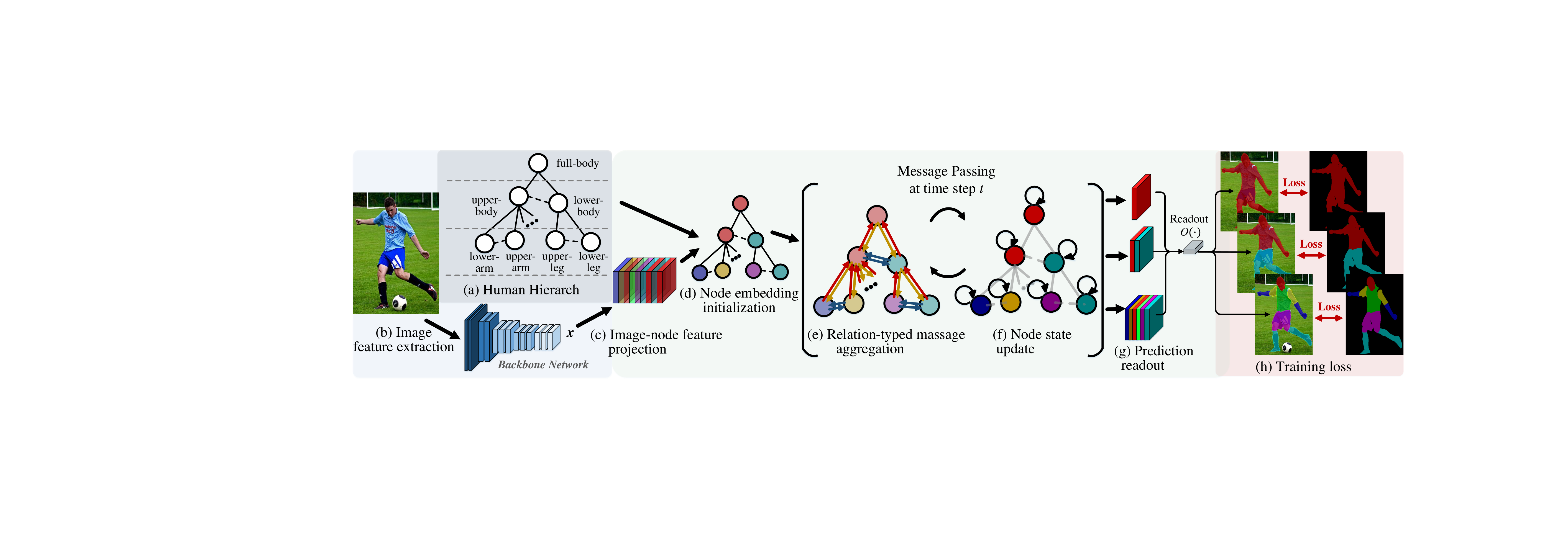}
      \put(-347,11.5){\scriptsize {(Eq.\!~\ref{eq:nodeini})}}
      \put(-235,11.5){\scriptsize {(Eq.\!~\ref{eq:message})}}
      \put(-173,12){\scriptsize {(Eq.\!~\ref{eq:update})}}
      \put(-112.5,4){\scriptsize {(Eq.\!~\ref{eq:readout})}}
      \put(-24,3.3){\scriptsize {(Eq.\!~\ref{eq:lossparsing})}}
      \put(-453,51){\scriptsize {\textcolor{ggray}{$\mathcal{V}_1$}}}
      \put(-453,73){\scriptsize {\textcolor{ggray}{$\mathcal{V}_2$}}}
      \put(-453,95){\scriptsize {\textcolor{ggray}{$\mathcal{V}_3$}}}
      \put(-388,39){\scriptsize {{~\!$\mathcal{G}$}}}
      \put(-414,28){\tiny {\textcolor{ggray}{$W_{_{\!}}\!\!\times_{_{\!}}\!\!H_{_{\!}}\!\!\times_{_{\!}}\!\!C$}}}
      \put(-372,60){\tiny {\textcolor{ggray}{$W_{_{\!}}\!\!\times_{_{\!}}\!\!H_{_{\!}}\!\!\times_{_{\!}}\!\!c_{_{\!}}\!\!\times_{_{\!}}\!\!|_{\!}\mathcal{V}_{\!}|$}}}
      \put(-372,28){\scriptsize $\{\textbf{\textit{h}}_v\!\}_{v\in\mathcal{V}}$}
      \put(-333,70){\scriptsize $\textbf{\textit{h}}_v^{(0)}$}
      \put(-279,55){\scriptsize $\textbf{\textit{h}}_v^{_{\!}(_{\!}t\!-\!1_{\!})}$}
      \put(-201,52){\scriptsize $\textbf{\textit{h}}_v^{_{\!}(_{\!}t_{\!})}$}
      \put(-197,72){\scriptsize $\textbf{\textit{m}}_v^{_{\!}(_{\!}t_{\!})}$}
      \put(-142,42.2){\scriptsize $\{_{\!}\textbf{\textit{h}}_{_{\!}v}^{_{\!}(\!t\!)}\!\}_{\!v_{\!}\in_{\!}\mathcal{V}_{_{\!}2}}$}
      \put(-136,98){\scriptsize $\{_{\!}\textbf{\textit{h}}_{_{\!}v}^{_{\!}(\!t\!)}\!\}_{\!v_{\!}\in_{\!}\mathcal{V}_{_{\!}3}}$}
      \put(-114,20){\scriptsize $\{_{\!}\textbf{\textit{h}}_{_{\!}v}^{_{\!}(\!t\!)}\!\}_{\!v_{\!}\in_{\!}\mathcal{V}_{_{\!}1}}$}
      \put(-70,34){\tiny {\textcolor{white}{$\hat{\mathbf{\mathcal{Y}}}^{_{\!}(_{\!}t_{\!})}_{_{\!}1}$}}}
      \put(-27,34){\tiny {\textcolor{white}{$\mathbf{\mathcal{Y}}_{_{\!}1}$}}}
      \put(-78,63){\tiny {\textcolor{white}{$\hat{\mathbf{\mathcal{Y}}}^{_{\!}(_{\!}t_{\!})}_{_{\!}2}$}}}
      \put(-35,63){\tiny {\textcolor{white}{$\mathbf{\mathcal{Y}}_{_{\!}2}$}}}
      \put(-86,92){\tiny {\textcolor{white}{$\hat{\mathbf{\mathcal{Y}}}^{_{\!}(_{\!}t_{\!})}_{_{\!}3}$}}}
      \put(-43,92){\tiny {\textcolor{white}{$\mathbf{\mathcal{Y}}_{_{\!}3}$}}}
\vspace{-5pt}
\captionsetup{font=small}
\caption{\small Illustration of our structured human parser for hierarchical human parsing during the training phase. The main components in the flowchart are marked by (a)-(h). Please refer to \S\ref{sec:approach} for  more details. Best viewed in color.}
\label{fig:model}
\vspace{-9pt}
\end{figure*}
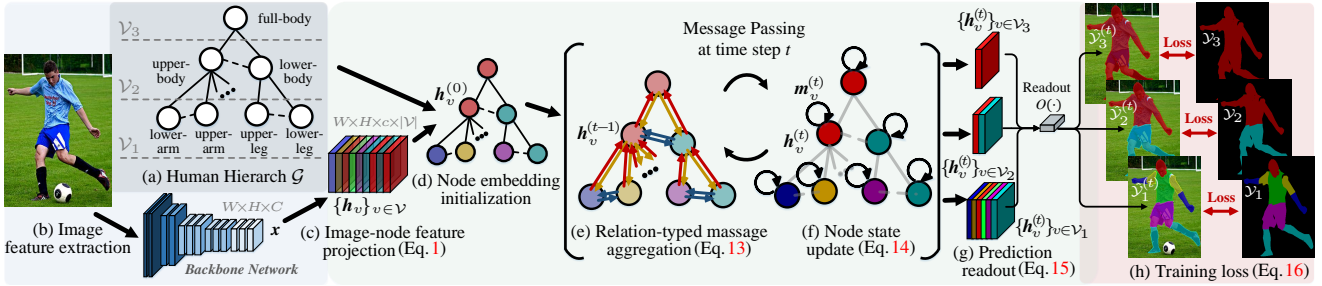

%From a broader perspective Other notable works. edge-typed graph
%relations
%Recently, researchers attempt to tackle this problem using deep learning. The most straightforward
%way is to directly predict Another stream
%Existing models [20, 21, 28, 8, 19] built on the encoder-decoder structure lack in considering the
%crucial shape and appearance misalignments, often leading to unsatisfying generated person images. modeling higher-level part-level structure layouts. Hence, their results
%suffer from various artifacts, blurry boundaries, missing clothing appearance when large geometric
%transformations are requested by the desirable poses, which are far from satisfaction.
%Consequent, most trained in an end-to-end fashion
% Our study is closely related to the following fields: semantic human part parsing, adaptive computation for neural networks, and hierarchical models in computer vision.

%Our model is generally fall in this category.
\vspace{-3pt}
\section{Our Approach}
\label{sec:approach}
\vspace{-3pt}
%-------------------------------------------------------------------------
\subsection{Problem Definition}
\vspace{-3pt}
Formally, we represent the human semantic structure as a directed,  hierarchical graph  $\mathcal{G}\!=\!(\mathcal{V}, \mathcal{E}, \mathcal{Y})$. As show in Fig.\!~\ref{fig:model}(a), the node set $\mathcal{V}\!=\!\cup_{l=1}^{3}\!\mathcal{V}_l$ represents human parts in three different semantic levels, including the leaf nodes $\mathcal{V}_1$ (\ie, the most fine-grained parts: \textit{head, arm, hand}, \etc)  which are typically considered in common human parsers, two middle-level nodes $\mathcal{V}_{2\!}\!=$\{\textit{upper-body, lower-body}\} and one root $\mathcal{V}_{3}\!=$\{\textit{full-body}\}\footnote{\scriptsize{As the classic settings of graph models, there is also a `dummy' node in $\mathcal{V}$, used for interpreting the background class. As it does not interact with other semantic human parts (nodes), we omit this node for concept clarity.}}. The edge set $\mathcal{E}\!\!\in\!\!\binom{\mathcal{V}}{2}$ represents the relations between human parts (nodes), \ie, the directed edge $e\!=\!(u, v)\!\in\!\mathcal{E}$ links node $u$ to $v_{\!}\!:\!u\!\!\rightarrow\!\!v$.  Each node $v$ and each edge $(u, v)$ are associated with feature vectors: $\textbf{\textit{h}}_v$ and $\textbf{\textit{h}}_{u, v}$, respectively. $y_{v\!}\!\in\!\mathcal{Y}$ indicates the groundtruth segmentation map of part (node) $v$ and the groundtruth maps $\mathcal{Y}$ are also organized in a hierarchical manner: $\mathcal{Y}\!=\!\cup_{l=1}^{3}\mathcal{Y}_l$.

Our human parser is trained in a graph learning scheme, using the full supervision from existing human parsing datasets. For a test sample, it is able to effectively infer the node and edge representations by reasoning human structures at the levels of individual parts and their relations, and iteratively fusing the information over the human structures.

\vspace{-2pt}
\subsection{Structured Human Parsing Network}\label{sec:3.2}
\vspace{-1pt}
\noindent\textbf{Node embedding:} As an initial step, a learnable projection function is used to map the input image representation into node (part) features, in order to obtain sufficient expressive power. Formally, let us denote the input image feature as $\textit{\textbf{x}}\!\in\!\mathbb{R}^{W\!\times\! H\!\times\!C\!}$, which comes from a DeepLabV3\!~\cite{chen2018deeplabv3plus}-like backbone network (\includegraphics[scale=0.08]{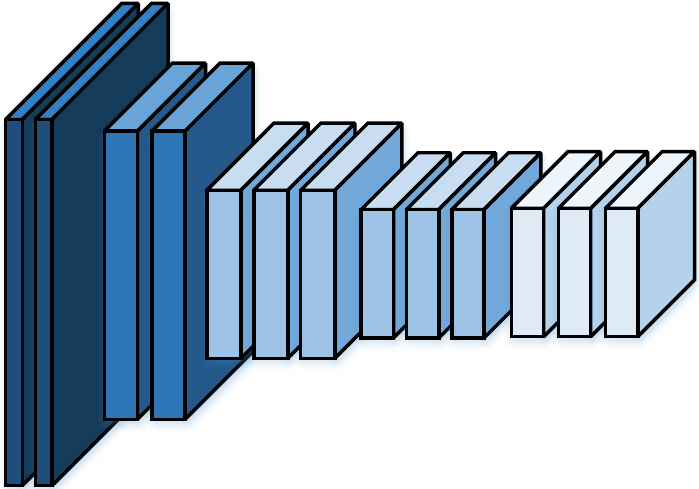} in Fig.\!~\ref{fig:model}(b)), and the projection function as $P\!\!:\!\mathbb{R}^{W\!\times\!H\!\times\!C}\!\!\mapsto\!\!\mathbb{R}^{W\!\times\!H\!\times c\times\!|\mathcal{V}|}$, where $|\mathcal{V}|$ indicates the number of nodes. The node embeddings $\{\textbf{\textit{h}}_v\!\!\in\!\!\mathbb{R}^{W\!\times\!H\!\times\!c}\}_{v\in\mathcal{V}}$ are initialized by (Fig.\!~\ref{fig:model}(d)):
\vspace{-2pt}
\begin{equation}
\begin{aligned}
\{\textbf{\textit{h}}_v\}_{v\in\mathcal{V}}\!=\!P(\textbf{\textit{x}}),
\end{aligned}
\label{eq:nodeini}
\vspace{-1pt}
\end{equation}
where each node embedding $\textbf{\textit{h}}_v$ is a ($W$\!, $H$\!, $c$)-dimensional tenor that encodes full spatial details (\includegraphics[scale=0.1]{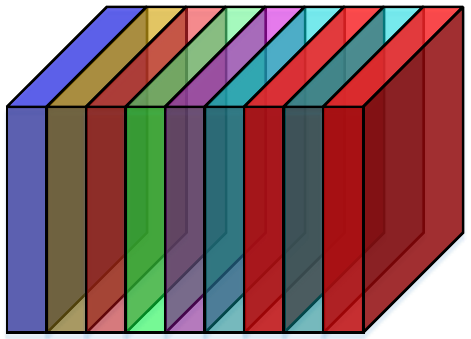} in Fig.\!~\ref{fig:model}(c)).

%Rather than thinking of question answering as a problem of learning a single function to map
%from questions and contexts to answers, it's perhaps useful
%to think of it as a highly-multitask learning setting, where
%each problem instance is associated with a novel task, and
%the identity of that task is expressed only noisily in language.

%Next, we will analyze the three kinds of human part relations, \ie, decomposition, composition, and dependency, and elaborate on our typed relation modeling strategy.

\noindent\textbf{Typed human part relation modeling:}
Basically, an edge embedding $\textbf{\textit{h}}_{u,v}$ captures the relations between nodes $u$ and $v$. Most previous structured human parsers~\cite{Zhu2018ProgressiveCH,wang2019CNIF} work in an edge-type-agnostic manner, \ie, a unified, shared relation network $R_{\!}\!:\!\mathbb{R}^{W\!\times\!H\!\times\!c\!}\!\times\!\mathbb{R}^{W\!\times\!H\!\times\!c\!}\!\mapsto\!\mathbb{R}^{W\!\times\!H\!\times\!c}$ is used to capture all the relations: $\textbf{\textit{h}}_{u,v\!}\!=\!R(\textbf{\textit{h}}_u, \textbf{\textit{h}}_v)$. Such a strategy may lose the discriminability of individual relation types and does not have an explicit bias towards modeling geometric and anatomical constraints.
In contrast, we formulate $\textbf{\textit{h}}_{u,v}$ in a relation-typed manner $R^r$:
\vspace{-3pt}
\begin{equation}\small
\begin{aligned}
\textbf{\textit{h}}_{u,v}=R^r(F^r(\textbf{\textit{h}}_u), \textbf{\textit{h}}_v),
\end{aligned}
\vspace{-2pt}
\label{eq:edge-specific}
\end{equation}
where $r_{\!}\!\!\in_{\!}\!\!\{\text{dec}, \text{com}, \text{dep}\}$. $F^{r\!}(\cdot)$ is an attention-based relation-adaption operation, which is used to enhance the original node embedding $\textbf{\textit{h}}_{u\!}$ by addressing geometric characteristics in relation $r$. The attention mechanism is favored here as it allows trainable and  flexible feature enhancement and explicitly encodes specific relation constraints. From the view of information diffusion mechanism in the graph theory~\cite{scarselli2008graph}, if there exists an edge ($u,v$) that links a starting node $u$ to a destination $v$, this indicates $v$ should receive incoming information (\ie, $\textbf{\textit{h}}_{u,v}$) from $u$. Thus, we use $F^{r\!}(\cdot)$ to make $\textbf{\textit{h}}_u$ better accommodate the target $v$. $R^r$ is edge-type specific, employing the more tractable feature $F^r(\textbf{\textit{h}}_u)$ in place of $\textbf{\textit{h}}_u$, so more expressive relation feature $\textbf{\textit{h}}_{u,v}$ for $v$ can be obtained and further benefit the final parsing results. In this way, we learn more sophisticated and impressive relation patterns within human
bodies.

%%%%%%%%%%%%%%%%%%% Figure 3%%%%%%%%%%%%%%%%%%%%%%
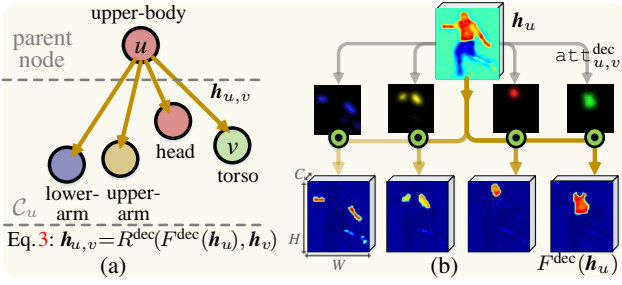
\begin{figure}[t]
%%tr = 0.006, ts = 0.008
  \centering
      \includegraphics[width=1 \linewidth]{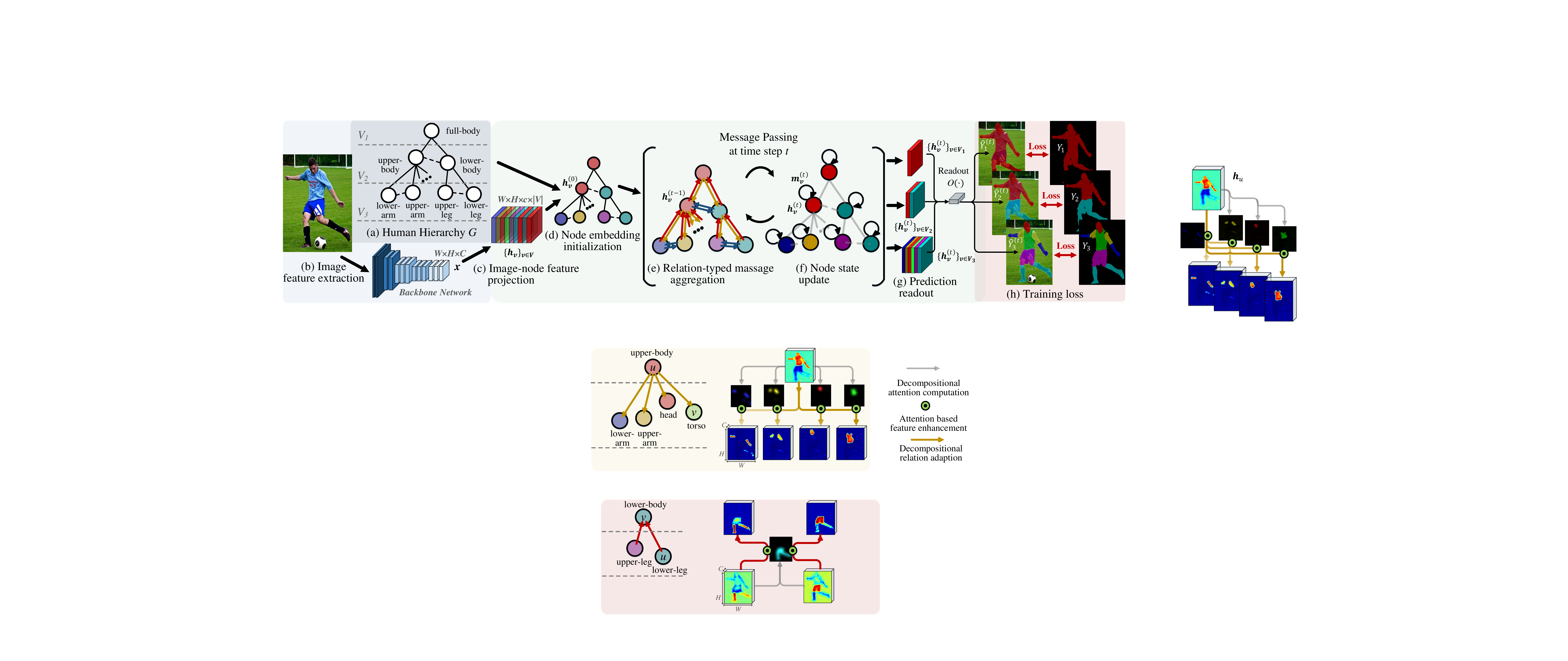}
      \put(-235,12){\footnotesize {Eq.\!~\ref{eq:decomposition}: $\textbf{\textit{h}}_{\!u,v\!}\!=_{\!}\!R^{\text{dec}\!}(_{\!}F^{\text{dec}}(\textbf{\textit{h}}_{u\!}), \textbf{\textit{h}}_{v\!})$}}
      \put(-44,95){\footnotesize {$\textbf{\textit{h}}_u$}}
      \put(-158,68){\footnotesize {$\textbf{\textit{h}}_{u,v}$}}
      \put(-35,3){\footnotesize {$F^{\text{dec}}(\textbf{\textit{h}}_u)$}}
      \put(-28,82){\footnotesize {$\texttt{att}_{u,v}^{\text{dec}}$}}
      \put(-200,2){\small {(a)}}
      \put(-75,2){\small {(b)}}
      \put(-233,24){\small {\textcolor{ggray}{$\mathcal{C}_u$}}}
      \put(-233,88){\small {\textcolor{ggray}{parent}}}
      \put(-233,80){\small {\textcolor{ggray}{~node}}}
\vspace{-6pt}
\captionsetup{font=small}
\caption{\small \textbf{Illustration of our decompositional relation modeling.} (a) Decompositional relations between the \textit{upper-body} node ($u$) and its constituents ($\mathcal{C}_u$). (b) With the decompositional attentions $\{\texttt{att}_{u,v}^{\text{dec}}(\textbf{\textit{h}}_u)\}_{v\in\mathcal{C}_u}$,  $F^{\text{dec}\!}$ learns how to `break down' the \textit{upper-body} node and generates more tractable features for its constituents. In the relation adapted feature $F^{\text{dec}}(\textbf{\textit{h}}_u)$, the responses from the background and other irrelevant parts are suppressed.}
\label{fig:decom_relation}
\vspace{-12pt}
\end{figure}

%set of discrete type variables are often used to

\noindent\textbf{1)} \textit{Decompositional relation modeling:} Decompositional relations (full line:\includegraphics[scale=0.16]{figs/de-com.pdf} in Fig.\!~\ref{fig:model}(a)) are represented by those vertical edges starting from parent nodes to corresponding child
nodes in the human hierarchy $\mathcal{G}$. For example, a parent node \textit{full-body} can be separated into \{\textit{upper-body}, \textit{lower-body}\}, and \textit{upper-body} can be decomposed into \{\textit{head}, \textit{torso}, \textit{upper-arm},  \textit{lower-arm}\}.
Formally, for a node $u$, let us denote its child node set as $\mathcal{C}_u$. Our decompositional relation network $R^{\text{dec}}$ aims to learn the rule for `breaking down' $u$ into its constituent parts $\mathcal{C}_u$ (Fig.\!~\ref{fig:decom_relation}):
%Decompositional relation has two properties: \textbf{(1)} the parent node and its child nodes are `compatible', \ie,  geometrically contains all of its constituents; and \textbf{(2)} the child nodes are mutually exclusive. Our decompositional relation function $R^{\text{dec}}$ is accordingly designed as:
\vspace{-3pt}
\begin{equation}\small
\begin{aligned}
\textbf{\textit{h}}_{u,v}\!&=\!R^{\text{dec}}(F^{\text{dec}}(\textbf{\textit{h}}_u), \textbf{\textit{h}}_v),~~~v\in\mathcal{C}_u,\\
F^{\text{dec}}(\textbf{\textit{h}}_u)\!&=\!\textbf{\textit{h}}_u\odot \verb"att"_{u,v}^{\text{dec}}(\textbf{\textit{h}}_u).
\end{aligned}
\label{eq:decomposition}
\vspace{-2pt}
\end{equation}
`$\odot$' indicates the attention-based feature enhancement operation, and $\verb"att"_{u,v\!}^{\text{dec}}(\textbf{\textit{h}}_{u\!})_{\!}\!\in_{\!}\![0,1]^{W\!\times\!H\!}$ produces an attention map.$_{\!}$ For$_{\!}$ each$_{\!}$ sub-node$_{\!}$ $v_{\!}\!\in_{\!}\!\mathcal{C}_{u\!}$ of $u$, $\verb"att"_{u,v\!}^{\text{dec}}(\textbf{\textit{h}}_{u\!})_{\!}$ is defined as:
\vspace*{-3pt}
\begin{equation}\small
\begin{aligned}
\!\!\!\!\verb"att"_{u,v}^{\text{dec}}(\textbf{\textit{h}}_{u\!}) \!= \! \text{PSM}([\phi^{\text{dec}\!}_v(\textbf{\textit{h}}_{u\!})]_{v\in\mathcal{C}_u\!})  \!=\!\frac{\exp(\phi^{\text{dec}}_v(\textbf{\textit{h}}_{u\!}))}{\Sigma_{v'\in\mathcal{C}_{u\!\!}}\exp(\phi^{\text{dec}\!}_{v'}(\textbf{\textit{h}}_{u\!}))},
\end{aligned}
\label{eq:decompositionatt}
\vspace*{3pt}
\end{equation}
where PSM($\cdot$) stands for \textit{pixel-wise soft-max}, `[$\cdot$]' represents the channel-wise concatenation, and $\phi^{\text{dec}}_v(\textbf{\textit{h}}_u)\!\in\!\mathbb{R}^{W\!\times\!H\!}$ computes a specific significance map for $v$. By making $\sum_{v\in\mathcal{C}_{u\!\!}}\verb"att"^{\text{dec}}_{u,v}\!=\!\textbf{1}$, $\{\verb"att"_{u,v}^{\text{dec}}(\textbf{\textit{h}}_u)\}_{v\in\mathcal{C}_u}$ forms a \textit{decompositional attention} mechanism, \ie, allocates disparate attentions over $\textbf{\textit{h}}_u$. To recap, the \textit{decompositional attention}, conditioned on $\textbf{\textit{h}}_u$, lets $u$ pass separate high-level information to different child nodes $\mathcal{C}_u$ (see Fig.\!~\ref{fig:decom_relation}(b)). Here $\verb"att"_{u,v}^{\text{dec}}(\cdot)$ is node-specific and separately learnt for the three entire nodes in $\mathcal{V}_2\cup\mathcal{V}_3$, namely \textit{full-body}, \textit{upper-body} and \textit{lower-body}. A subscript $_{u,v}$ is added to address this point.
In addition, for each parent node $u$, the groundtruth maps $\mathcal{Y}_{\mathcal{C}_{u}\!}\!=\!_{\!}\{y_v\}_{v\in\mathcal{C}_u\!}\!\in\!\{0,1\}^{W\!\times\!H\!\times\!|\mathcal{C}_{u}|}$ of all the child nodes $\mathcal{C}_{u}$ can be used as supervision signals to train its \textit{decompositional attention} $\{\verb"att"_{u,v}^{\text{dec}}(\textbf{\textit{h}}_u)\}_{v\in\mathcal{C}_u}\!\in\![0,1]^{W\!\times\!H\times\!|\mathcal{C}_{u}|}$:
\vspace{-3pt}
\begin{equation}\small
\begin{aligned}
\mathcal{L}_{\text{dec}} = \sum\nolimits_{u\in\mathcal{V}_2\cup\mathcal{V}_3}\mathcal{L}_{\text{CE}}\big(\{\verb"att"_{u,v}^{\text{dec}}(\textbf{\textit{h}}_u)\}_{v\in\mathcal{C}_u}, \mathcal{Y}_{\mathcal{C}_{u}}\big),
\end{aligned}
\label{eq:decomloss}
\vspace{-1pt}
\end{equation}
where $\mathcal{L}_{\text{CE}}$ represents the standard cross-entropy loss.

%In this way, the parent node $u$ passes separate high-level information to different child nodes.

%The term $A(\textbf{\textit{h}}_u)$ is used to addresses the first property. It enforces the compatibility between $u$ and $\mathcal{C}_u$, \ie, lets the decompositional attentions only focus on the region within $u$, while suppresses the responses from other regions.

%Concerning $F^{\text{dec}}_{u,v}$, please note that it is node-specific and separately learnt for the three entire nodes, namely \textit{full-body}, \textit{upper-body} and \textit{lower-body}. A subscript $_{u,v}$ is added to address this point.

%Remark

\begin{figure}[t]
%%tr = 0.006, ts = 0.008
  \centering
      \includegraphics[width=1 \linewidth]{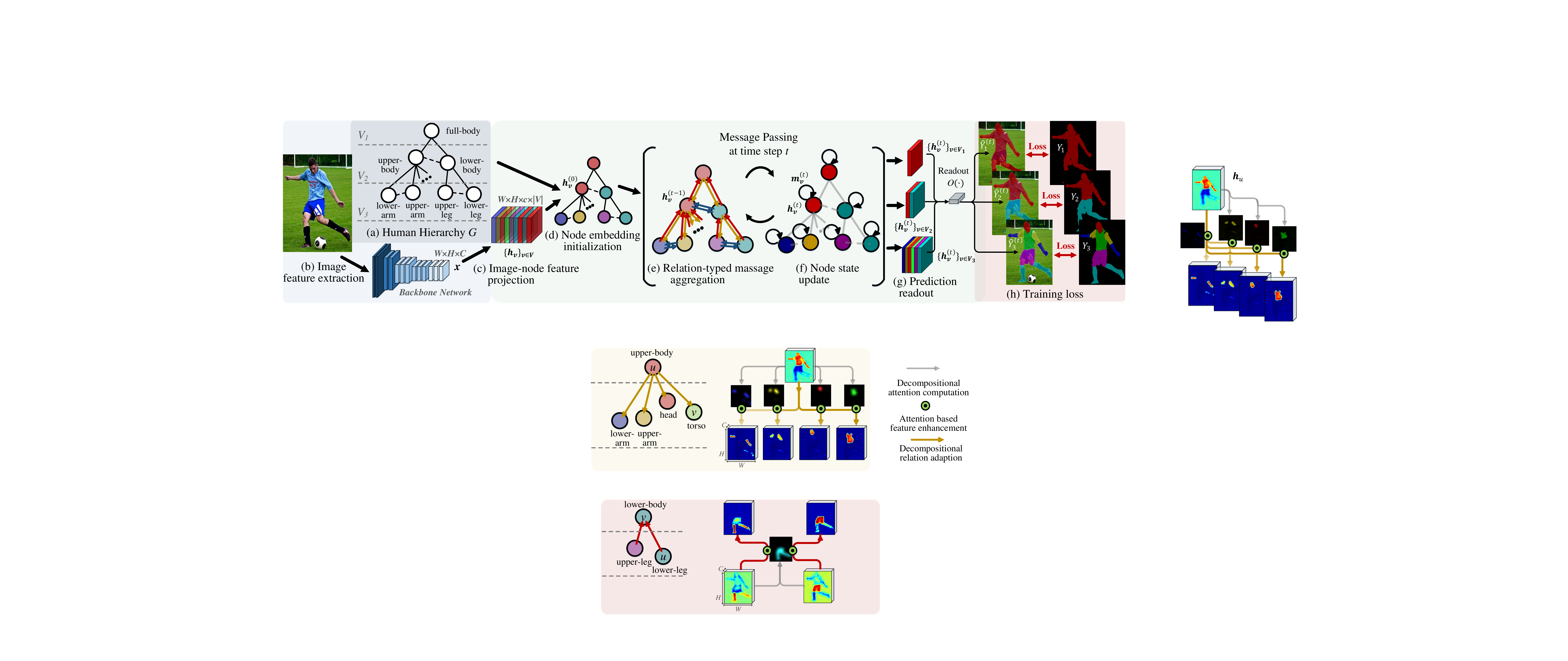}
      \put(-235,21){\footnotesize {Eq.\!~\ref{eq:composition}:}}
      \put(-235,12){\footnotesize { $\textbf{\textit{h}}_{\!u,v\!}\!=_{\!}\!R^{\text{com}\!}(_{\!}F^{\text{com}}(\textbf{\textit{h}}_{u\!}), \textbf{\textit{h}}_{v\!})$}}
      \put(-34,24){\footnotesize {$\textbf{\textit{h}}_u$}}
      \put(-148,24){\footnotesize {$\textbf{\textit{h}}_{u'\!}$}}
      \put(-188,63){\footnotesize {$\textbf{\textit{h}}_{u,v}$}}
      \put(-212,52){\small {$u'$}}
      \put(-34,80){\footnotesize {$F^{\text{com}}(\textbf{\textit{h}}_u)$}}
      \put(-168,80){\footnotesize {$F^{\text{com}}(\textbf{\textit{h}}_{u'\!})$}}
      \put(-97,72){\footnotesize {$\texttt{att}_{v}^{\text{com}}$}}
      \put(-200,2){\small {(a)}}
      \put(-75,2){\small {(b)}}
      \put(-99,15){\small {$[\textbf{\textit{h}}_{u'\!}, \textbf{\textit{h}}_{u}]$}}
      \put(-235,35){\small {\textcolor{ggray}{$\mathcal{C}_v$}}}
      \put(-235,82){\footnotesize {\textcolor{ggray}{parent}}}
      \put(-235,74){\footnotesize {\textcolor{ggray}{~node}}}
\vspace{-6pt}
\captionsetup{font=small}
\caption{\small \textbf{Illustration of our compositional relation modeling.} (a) Compositional relations between the \textit{lower-body} node ($v$) and its constituents ($\mathcal{C}_v$). (b) The compositional attention $\texttt{att}_{v}^{\text{com}}([\textbf{\textit{h}}_{u'}, \textbf{\textit{h}}_{u}])$ gathers information from all the constituents $\mathcal{C}_v$ and lets $F^{\text{com}\!}$ enhance all the \textit{lower-body} related features of $\mathcal{C}_v$.}
\label{fig:com_relation}
\vspace{-12pt}
\end{figure}
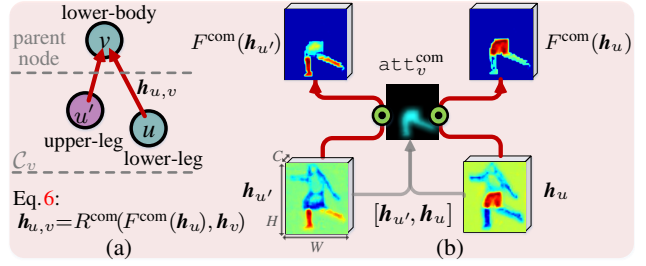

\noindent\textbf{2)} \textit{Compositional relation modeling:} In the human hierarchy $\mathcal{G}$, compositional relations are represented by vertical, downward edges. To address this type of relations, we design a compositional relation network $R^{\text{com}}$ as (Fig.\!~\ref{fig:com_relation}):
\vspace{-2pt}
\begin{equation}\small
\begin{aligned}
\textbf{\textit{h}}_{u,v}\!&=\!R^{\text{com}}(F^{\text{com}}(\textbf{\textit{h}}_{u}), \textbf{\textit{h}}_{v}),~~~u\in\mathcal{C}_v,\\
F^{\text{com}}(\textbf{\textit{h}}_{u})\!&=\!\textbf{\textit{h}}_{u}\!\odot\!\verb"att"_{v}^{\text{com}}([\textbf{\textit{h}}_{u'}]_{u'\in \mathcal{C}_v}).
\end{aligned}
\label{eq:composition}
\vspace{-1pt}
\end{equation}
Here $\verb"att"_{v}^{\text{com}\!}\!\!:\!\mathbb{R}^{W\!\times\!H\!\times\!c\!\times\!|\mathcal{C}_v|\!}\!\mapsto_{\!}\![0,1]^{W\!\times\!H\!}$
is a \textit{compositional attention}, implemented by a $1\!\times\!1$ convolutional layer. The rationale behind such a design is that, for a parent node $v$, $\verb"att"_{v}^{\text{com}}$ gathers statistics of all the child nodes $\mathcal{C}_v$ and is used to enhance each sub-node feature $\textbf{\textit{h}}_{u}$. As $\verb"att"^{\text{com}}_v$ is compositional in nature, its enhanced feature $F^{\text{com}}(\textbf{\textit{h}}_{u})$ is more `friendly' to the parent node $v$, compared to $\textbf{\textit{h}}_{u}$. Thus, $R^{\text{com}}$ is able to generate more expressive relation features by considering compositional structures (see Fig.\!~\ref{fig:com_relation}(b)).

%%%%%%%%%%%%%%%%%%% Figure 4%%%%%%%%%%%%%%%%%%%%%%

For each parent node $v\!\in\!\mathcal{V}_2\cup\mathcal{V}_3$, with its groundtruth map $y_v\!\in\!\{0,1\}^{W\!\times\!H}$, the \textit{compositional attention} for all its child nodes $\mathcal{C}_v$ is trained by minimizing the following loss:
\vspace{-2pt}
\begin{equation}\small
\begin{aligned}
\mathcal{L}_{\text{com}} = \sum\nolimits_{v\in\mathcal{V}_2\cup\mathcal{V}_3}\mathcal{L}_{\text{CE}}\big(\verb"att"_{v}^{\text{com}}([\textbf{\textit{h}}_{u'}]_{u'\in \mathcal{C}_v}), y_v\big).
\end{aligned}
\label{eq:comloss}
\vspace{-1pt}
\end{equation}

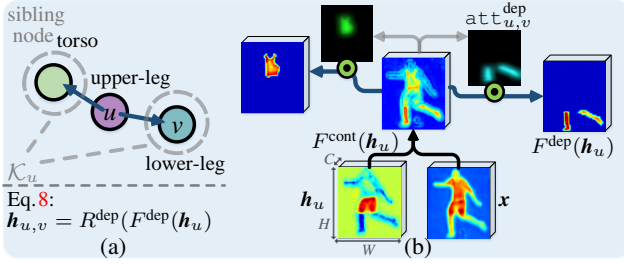
\begin{figure}[t]
%%tr = 0.006, ts = 0.008
  \centering
      \includegraphics[width=1 \linewidth]{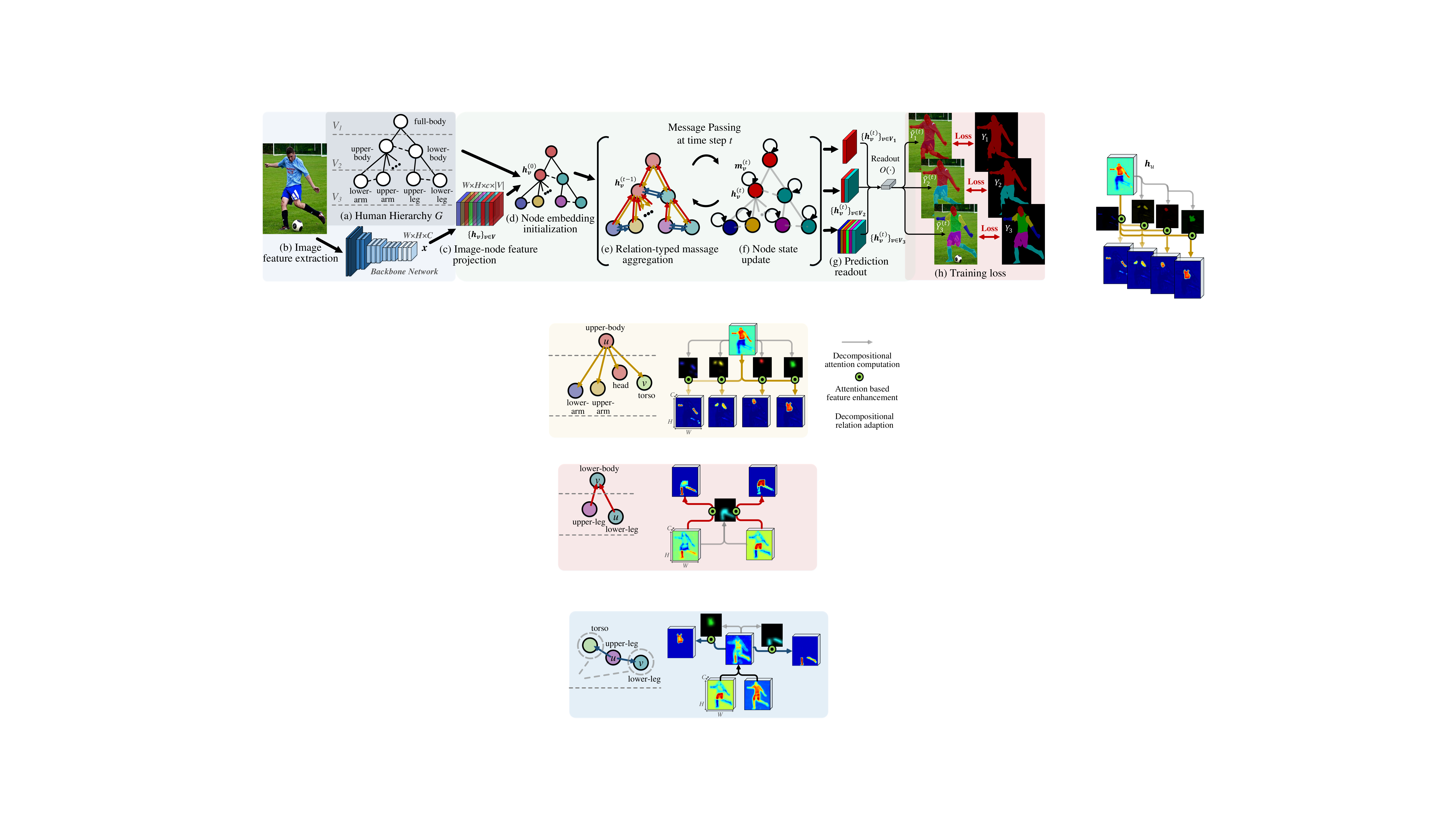}
      \put(-235,20){\footnotesize {Eq.\!~\ref{eq:dependency}:}}
      \put(-235,12){\footnotesize {$\textbf{\textit{h}}_{u,v} = R^{\text{dep}}(F^{\text{dep}}(\textbf{\textit{h}}_u)$}}
      \put(-124,20){\footnotesize {$\textbf{\textit{h}}_u$}}
      \put(-49,20){\footnotesize {$\textbf{\textit{x}}$}}
      \put(-37,40){\footnotesize {$F^{\text{dep}}(\textbf{\textit{h}}_u)$}}
      \put(-120,42){\footnotesize {$F^{\text{cont}}(\textbf{\textit{h}}_u)$}}
      \put(-62,89){\footnotesize {$\texttt{att}_{u,v}^{\text{dep}}$}}
      \put(-200,2){\small {(a)}}
      \put(-85,2){\small {(b)}}
      \put(-235,30){\small {\textcolor{ggray}{$\mathcal{K}_u$}}}
      \put(-235,90){\footnotesize {\textcolor{ggray}{sibling}}}
      \put(-235,82){\footnotesize {\textcolor{ggray}{~node}}}
\vspace{-6pt}
\captionsetup{font=small}
\caption{\small \textbf{\!\!\!Illustration of our dependency relation modeling.} (a) Dependency relations between the \textit{upper-body} node ($u$) and its siblings ($\mathcal{K}_u$). (b) The dependency attention $\{_{\!}\texttt{att}_{u,v\!}^{\text{dep}}\big(_{\!}F^{\text{cont\!}}(\textbf{\textit{h}}_{u\!})_{\!}\big)_{\!}\}_{\!v\in\mathcal{K}_u\!}$, derived from $u$'s contextual information $F^{\text{cont}}(\textbf{\textit{h}}_u)$, gives separate importance for different siblings $\mathcal{K}_u$.}
\label{fig:dep_relation}
\vspace{-12pt}
\end{figure}

\noindent\textbf{3)} \textit{Dependency relation modeling:} In $\mathcal{G}$, dependency relations are represented as horizontal edges (dashed line:\includegraphics[scale=0.16]{figs/depen.pdf} in Fig.\!~\ref{fig:model}(a)), describing pairwise, kinematic connections between human parts, such as (\textit{head}, \textit{torso}), (\textit{upper-leg}, \textit{lower-leg}), \etc.
% it aggregates information from input
%score maps2 on a spatially local support to predict output score maps.
%
%spatially local information summarizationSegmenting a part with regard to its surrounding parts is super helpful for reducing local ambiguities.
Two kinematically connected human parts are spatially adjacent, and their dependency relation essentially addresses the context information. For a node $u$, with its kinematically connected siblings $\mathcal{K}_u$, a dependency relation network  $R^{\text{dep}}$ is designed as (Fig.~\ref{fig:dep_relation}):
%The main idea is that, given a node u, a dependency network is learned to take $h_u$ as input, and generate an edge embedding $h_{u,v}$ that accounts above dependency constrains.
\vspace{-2pt}
\begin{equation}\small
\begin{aligned}
\textbf{\textit{h}}_{u,v} &= R^{\text{dep}}(F^{\text{dep}}(\textbf{\textit{h}}_u), \textbf{\textit{h}}_v),~~~~v\in\mathcal{K}_u,\\
F^{dep}(\textbf{\textit{h}}_u)&=F^{\text{cont}}(\textbf{\textit{h}}_u)\!\odot\!\verb"att"^{\text{dep}}_{u,v}\big(F^{\text{cont}}(\textbf{\textit{h}}_u)\big),
\end{aligned}
\label{eq:dependency}
\vspace{-1pt}
\end{equation}
where $F^{\text{cont}}(\textbf{\textit{h}}_{u\!})\!\in\!\mathbb{R}^{W\!\times\!H\!\times\! c\!}$ is used to extract the context of $u$, and $\verb"att"^{\text{dep}}_{u,v}\big(F^{\text{cont}}(\textbf{\textit{h}}_u)\big)\!\in\![0,1]^{W\!\times\!H\!}$ is a \textit{dependency attention} that produces an attention for each sibling node $v$, conditioned on $u$'s context $F^{\text{cont}}(\textbf{\textit{h}}_u)$. Specifically, inspired by the non-local self-attention~\cite{vaswani2017attention,wang2018non}, the \textit{context extraction} module $F^{\text{cont}}$ is designed as:
\vspace{-2pt}
\begin{equation}\small
\begin{aligned}
F^{\text{cont}}(\textbf{\textit{h}}_u) &= \rho(\textbf{\textit{x}}\textbf{\textit{A}}^{\!\top})~\in \mathbb{R}^{W\!\times\!H\!\times\!c},\\
\textbf{\textit{A}}&={\textbf{\textit{h}}'_u}^{\!\!\top} \textbf{\textit{W}}\textbf{\textit{x}}'\in \mathbb{R}^{(W\!H)\times (W\!H)},
\end{aligned}
\label{eq:contex}
\vspace{-1pt}
\end{equation}
where $\textbf{\textit{h}}'_{u\!\!}\!\in_{\!}\!\mathbb{R}^{\!(c+8)\!\times\!(W\!H)\!}$ and $\textbf{\textit{x}}'^{\!}\!\!\in_{\!}\!\mathbb{R}^{(C+8)\!\times\!(W\!H)}$ are node (part) and image representations augmented with spatial information, respectively, flattened into matrix formats. The last eight channels of $\textbf{\textit{h}}'_u$ and $\textbf{\textit{x}}'$  encode spatial coordinate information~\cite{hu2016segmentation}, where the first
six dimensions are the normalized horizontal and vertical positions, and the last two dimensions
are the normalized width and height information of the feature, 1/$W_{\!}$ and 1/$H_{\!}$. %$\varphi(\cdot)\!\!:\!\mathbb{r}_{\geq0}^{h\!\times\!w\!\times\!(c+8)\!}\!\mapsto\!\mathbb{r}_{\geq0}^{h\!\times\!w\!\times\!c\!}$ and $\rho(\cdot)\!\!:\!\mathbb{r}_{\geq0}^{h\!\times\!w\!\times\!(c+8)\!}\!\mapsto\!\mathbb{r}_{\geq0}^{h\!\times\!w\!\times\!c\!}$  are $1\!\times\!1$ convolution based linear embedding functions that map the input features to a context space.
$\textbf{\textit{W}}_{\!}\!\in_{\!}\!\mathbb{R}^{(c+8)\times(C+8)\!}$ is learned as a linear transformation based node-to-context projection function. %in a hyperdimensional real space
The node feature $\textbf{\textit{h}}'_u$, used as a \textit{query} term,  retrieves the \textit{reference} image feature $\textbf{\textit{x}}'$ for its context information. As a result, the affinity matrix $\textbf{\textit{A}}$ stores the attention weight between the query and reference  at a certain spatial location, accounting for both visual and spatial information. Then, $u$'s context is collected as a weighted sum of the original image feature $\textbf{\textit{x}}$ with column-wise normalized weight matrix $\textbf{\textit{A}}^{\!\!\top\!\!}$: $_{\!}\textbf{\textit{x}}\textbf{\textit{A}}^{\!\!\top\!\!}\!\in_{\!}\!\mathbb{R}^{C\!\times\!(W\!H)\!}$.  A $1\!\times\!1$ convolution based linear embedding function $\rho\!:\!\mathbb{R}^{W\!\times\!H\!\times\!C\!\!}\!\mapsto_{\!\!}\!\mathbb{R}^{W\!\times\!H\!\times\!c\!}$  is applied for feature dimension compression, \ie, to make the channel dimensions
of different edge embeddings consistent.

%%%%%%%%%%%%%%%%%%% Figure 5%%%%%%%%%%%%%%%%%%%%%%

For each sibling node $v\!\in\!\mathcal{K}_u$ of $u$, $\verb"att"^{\text{dep}}_{u,v}$ is defined as:
\vspace*{-6pt}
\begin{equation}\small
\begin{aligned}
\verb"att"_{u,v}^{\text{dep}}\big(F^{\text{cont}}(\textbf{\textit{h}}_u)\big) &= \text{PSM}\big([\phi^{\text{dep}}_v(\textbf{\textit{h}}_u)]_{v\in\mathcal{K}_u}\big).
\end{aligned}
\vspace*{0pt}
\end{equation}
Here $\phi^{\text{dep}}_v(\cdot)\!\in\!\mathbb{R}^{W\!\times\!H\!}$ gives an importance map for $v$, using a $1\!\times\!1$ convolutional layer. Through the \textit{pixel-wise soft-max} operation PSM($\cdot$), we enforce $\sum_{v\in\mathcal{K}_{u\!\!}}\verb"att"^{\text{dep}}_{u,v}\!=\!\textbf{1}$, leading to a \textit{dependency attention} mechanism which assigns exclusive attentions over $F^{\text{cont}}(\textbf{\textit{h}}_u)$, for the corresponding sibling nodes $\mathcal{K}_u$. Such a \textit{dependency attention} is learned via:
\vspace{-2pt}
\begin{equation}\small
\begin{aligned}
\mathcal{L}_{\text{dep}} = \sum\nolimits_{u\in\mathcal{V}_1\cup\mathcal{V}_2}\mathcal{L}_{\text{CE}}\big(\{\verb"att"_{u,v}^{\text{dep}}(\textbf{\textit{h}}_u)\}_{v\in\mathcal{K}_u}, \mathcal{Y}_{\mathcal{K}_{u}}\big),
\end{aligned}
\label{eq:dependencyloss}
\vspace{-1pt}
\end{equation}
where $\mathcal{Y}_{\mathcal{K}_{u\!\!}}\!\in\![0,1]^{W\!\times\!H\!\times\!|\mathcal{K}_{u}|}$ stands for the groundtruth maps $\{y_v\}_{v\in\mathcal{K}_u}$  of all the sibling nodes $\mathcal{K}_{u}$ of $u$.

\noindent\textbf{Iterative inference over human hierarchy:} Human bodies present a hierarchical structure. According to graph theory, approximate inference algorithms should be
used for such a loopy structure $\mathcal{G}$. However, previous structured human parsers directly produce the final node representation $\textbf{\textit{h}}_v$ by either simply accounting for the information from the parent node $u_{\!}$~\cite{Zhu2018ProgressiveCH}: $\textbf{\textit{h}}_v\!\!\leftarrow\!\!R(\textbf{\textit{h}}_u, \textbf{\textit{h}}_v)$, where $v\!\in\!\mathcal{C}_u$; or from its neighbors $\mathcal{N}_{v\!}$~\cite{wang2019CNIF}: $\!\textbf{\textit{h}}_v\!\!\leftarrow\!\!\sum_{u\in\mathcal{N}_v\!}R(\textbf{\textit{h}}_u, \textbf{\textit{h}}_v)$. They ignore the fact that, in such a structured setting, information is organized in a complex system. Iterative algorithms offer a more favorable solution, \ie,  the node representation should be updated iteratively by aggregating the messages from
its neighbors; after several iterations, the representation can approximate the optimal results~\cite{scarselli2008graph}. %In graph theory parlance, the iterative algorithm has two phases, a message passing phase and a readout phase. First, the parametric message
In graph theory parlance, the iterative algorithm can be achieved by a parametric message
passing process, which is defined in terms of a message function $M$ and node update function $U$, and runs $T$ steps. For each node $v$, the message passing process recursively collects information (messages) $\textbf{\textit{m}}_v$ from the neighbors $\mathcal{N}_v$ to enrich the node embedding $\textbf{\textit{h}}_v$:
\vspace{-5pt}
\begin{equation}\small
\begin{aligned}
\textbf{\textit{m}}_v^{(t)} &= \sum\nolimits_{u\in\mathcal{N}_v}M(\textbf{\textit{h}}_u^{(t-1)}, \textbf{\textit{h}}_v^{(t-1)}),\\
\textbf{\textit{h}}_v^{(t)} &= U(\textbf{\textit{h}}_v^{(t-1)}, \textbf{\textit{m}}_v^{(t)}),
\end{aligned}
\vspace{-7pt}
\end{equation}
where $\textbf{\textit{h}}_v^{(t)}$ stands for $v$'s state in the $t$-th iteration. Recurrent neural networks are typically used to address the iterative nature of the update function $U$.

Inspired by previous message passing algorithms, our iterative algorithm is designed as (Fig.\!~\ref{fig:model}(e)-(f)):
\vspace{-4pt}
\begin{equation}\small
\begin{aligned}
\textbf{\textit{m}}_v^{(t)} \!=\!\underbrace{\sum\nolimits_{u\in\mathcal{P}_v}\!\textbf{\textit{h}}^{(t-1)}_{u,v}}_{\text{decomposition}}\!+\!\underbrace{\sum\nolimits_{u\in\mathcal{C}_v}\!\textbf{\textit{h}}^{(t-1)}_{u,v}}_{\text{composition}}\!+\!\underbrace{\sum\nolimits_{u\in\mathcal{K}_v}\!\textbf{\textit{h}}^{(t-1)}_{u,v}}_{\text{dependency}}
,\\
\end{aligned}
\label{eq:message}
\vspace{-4pt}
\end{equation}
\vspace{-6pt}
\begin{equation}\small
\begin{aligned}
\!\!\!\!\!\!\!\!\!\!\!\!\!\!\!\!\!\!\!\!\!\!\!\!\!\!\!\!\!\!\!\!\!\!\!\!\!\!\!\!\!\!\!\!\!\!\!\!\!\!\!\!\!\!\!\!\!\!\!\!\!\!\!\!\!\!\!\!\!\!\!\!\textbf{\textit{h}}_v^{(t)} \!=\! U_{\text{convGRU}}(\textbf{\textit{h}}_v^{(t-1)}\!, \textbf{\textit{m}}_v^{(t)}),
\end{aligned}
\label{eq:update}
\vspace{-2pt}
\end{equation}
where the initial state $\textbf{\textit{h}}_v^{(0)}$ is obtained by Eq.\!~\ref{eq:nodeini}. Here, the message aggregation step (Eq.\!~\ref{eq:message}) is achieved by per-edge relation function terms, \ie, node $v$ updates its state $\textbf{\textit{h}}_v$ by absorbing all the incoming information along different relations. As for the update function $U_{\!}$ in Eq.\!~\ref{eq:update}, we use a convGRU~\cite{xingjian2015convolutional}, which replaces the fully-connected units in the original MLP-based GRU with convolution operations, to describe its repeated activation behavior and address the pixel-wise nature of human parsing, simultaneously. Compared to previous parsers, which are typically based on \textit{feed-forward} architectures, our massage-passing inference essentially provides a \textit{feed-back} mechanism, encouraging effective reasoning over the cyclic human hierarchy $\mathcal{G}$.

%the solutions of the algorithms are characterized by a set of steady-state conditions

\noindent\textbf{Loss function:} In each step $t$, to obtain the predictions $\hat{\mathcal{Y}}^{(t)\!\!}_l\!=\!\!\{\hat{y}_v^{(t)\!\!\!}\!\in\!_{\!}[0, 1]^{W\!\times\!H}\}_{v\in\mathcal{V}_{l}\!}$ of the $l$-th layer nodes $\mathcal{V}_l$, we apply a convolutional readout function $O\!\!:\! \mathbb{R}^{W\!\times\!H\!\times\!c\!}\!\mapsto_{\!}\!\mathbb{R}^{W\!\times\!H}$ over $\{\textbf{\textit{h}}_v^{(t)}\}_{v\in\mathcal{V}}$ (\includegraphics[scale=0.24]{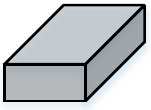} in Fig.\!~\ref{fig:model}(g)), and \textit{pixel-wise soft-max} (PSM) for normalization:
\vspace{-4pt}
\begin{equation}\small
\begin{aligned}
\hat{\mathcal{Y}}^{(t)}_l=\{\hat{y}_v^{(t)}\}_{v\in\mathcal{V}_l}=\text{PSM}\big([O(\textbf{\textit{h}}_v^{(t)})]_{v\in\mathcal{V}_l}\big).
\end{aligned}
\label{eq:readout}
\vspace{-6pt}
\end{equation}

Given the hierarchical human parsing results $\{\hat{\mathcal{Y}}^{(t)}_l\}^3_{l=1}$ and corresponding groundtruths $\{\mathcal{Y}_l\}^3_{l=1}$,
the learning task in the iterative inference can be posed as the minimization of the following loss (Fig.\!~\ref{fig:model}(h)):
\vspace{-5pt}
\begin{equation}\small
\begin{aligned}
\mathcal{L}_{\text{parsing}}^{(t)} = \sum\nolimits_{l=1}^{3}\mathcal{L}^{(t)}_{\text{CE}}(\hat{\mathcal{Y}}^{(t)}_l, \mathcal{Y}_l).
\end{aligned}
\label{eq:lossparsing}
\vspace{-4pt}
\end{equation}
Considering$_{\!}$ Eqs.\!~\ref{eq:decomloss},\!~\ref{eq:comloss},\!~\ref{eq:dependencyloss},\!~and\!~\ref{eq:lossparsing},$_{\!}$ the$_{\!}$ overall$_{\!}$ loss$_{\!}$ is$_{\!}$ defined$_{\!}$ as:
\vspace{-4pt}
\begin{equation}\small
\begin{aligned}
\mathcal{L} = \sum\nolimits_{t=1}^{T}\big(\mathcal{L} _{\text{parsing}}^{(t)}+\alpha(\mathcal{L}_{\text{com}}^{(t)}+\mathcal{L}_{\text{dec}}^{(t)}+\mathcal{L}_{\text{dep}}^{(t)})\big),
\end{aligned}
\vspace{-0pt}
\label{eq:lossall}
\end{equation}
where the coefficient $\alpha$ is empirically set as $0.1$. We set the total inference time $T\!=\!2$ and study how the performance changes with the number of inference iterations in \S\ref{sec:ablation}. %A pseudo-code description for our structured parser is provided in the supplementary material.
%a quadratic cost function
% standard supervised ultimately
%learning objective evaluated at node  applying the final nonlinear-
%ity (usually a softmax
%
%Finally, the
%output is computed wit. maps from node representations to an output   for each
\vspace{-3pt}
\subsection{Implementation Details}
\label{sec:network}
\vspace{-2pt}
\noindent\textbf{Node embedding:} A DeepLabV3 network\!~\cite{chen2018deeplabv3plus} serves as the backbone architecture, resulting in a 256-channel image representation whose spatial dimensions are 1/8 of the input image. The projection function $P\!\!:\!\mathbb{R}^{W\!\times\!H\!\times\!C\!}\!\mapsto\!\mathbb{R}^{W\!\times\!H\!\times c\times\!|\mathcal{V}|}$ in Eq.~\!\ref{eq:nodeini} is implemented by a $3\!\times\!3$ convolutional layer with ReLU nonlinearity, where $C_{\!}\!=_{\!}\!256$ and $|\mathcal{V}|$ (\ie, the number of nodes) is set according to the settings in different human parsing datasets. We set the channel size of node features $c\!=\!64$ to maintain high computational efficiency.

\noindent\textbf{Relation networks:}$_{\!}$ Each typed relation network $R^{r\!}$ in Eq.~\!\ref{eq:edge-specific} concatenates the relation-adapted feature $F^r(\textbf{\textit{h}}_u)$ from the source node $u$ and the destination node $v$'s feature $\textbf{\textit{h}}_v$ as the input, and outputs the relation representations: $\textbf{\textit{h}}_{u,v\!}\!=_{\!}\!R^{r\!}([F^{r\!}(\textbf{\textit{h}}_{u\!}), \textbf{\textit{h}}_v])$. $_{\!}R^{r\!}\!\!:_{\!}\!\mathbb{R}^{W\!\times\!H\!\times\!2c\!}\!\mapsto_{\!}\!\mathbb{R}^{W\!\times\!H\!\times\!c\!}$ is implemented by a $3\!\times\!3$ convolutional layer with ReLU nonlinearity.

\noindent\textbf{Iterative inference:} In Eq.~\!\ref{eq:update}, the update function $U_{\text{convGRU}}$ is implemented by a convolutional GRU with $3\!\times\!3$ convolution kernels.
The readout function $O$ in Eq.~\!\ref{eq:readout} applies a $1\!\times\!1$  convolution operation on the feature-prediction projection. In addition, before sending a node feature $\textbf{\textit{h}}_v^{(t)}$ into $O$, we use a light-weight decoder (built using a principle of upsampling the node feature and merging it with the low-level feature of the backbone network) that outputs the segmentation mask with 1/4 the spatial resolution of the input image.

As seen, all the units of our parser are built on convolution operations, leading to spatial information preservation.

\vspace{-3pt}
%------------------------------------------------------------------------
\section{Experiments}
\label{sec:exp}%
\vspace{-3pt}
%Herein, we describe our experimental settings (\S\ref{sec:dataset}), report quantitative results comparing to several state-of-the-arts on five datasets ($\sim$20K test images in total, \S\ref{sec:qresults1}), show qualitative results (\S\ref{sec:qresults2}), and study the impact of different components of our model (\S\ref{sec:ablation}). All the visual results shown in this section are drawn from the test sets. More quantitative and qualitative results are provided in the supplementary material.

\subsection{Experimental Settings}
\label{sec:dataset}
\vspace{-3pt}
\noindent\textbf{Datasets:}\footnote{As the datasets provide different human part labels, we make proper modifications of our human hierarchy. For some labels that do not deliver human structures, such as \textit{hat}, \textit{sun-glasses}, we treat them as isolate nodes.} Five standard benchmark datasets~\cite{gong2017look,xia2017joint,luo2013pedestrian,liang2015deep,Luo_2018_TGPnet}
are used for performance evaluation. LIP~\cite{gong2017look} contains 50,462 single-person images, which are collected from realistic scenarios and divided into 30,462 images for training, 10,000 for validation and 10,000 for test. The pixel-wise annotations cover 19 human part categories (\eg, \textit{face}, \textit{left-/right-arms}, \textit{left-/right-legs}, \etc). PASCAL-Person-Part~\cite{xia2017joint} includes 3,533 multi-person images with challenging poses and viewpoints. Each image is pixel-wise annotated with six classes (\ie, \textit{head, torso, upper-/lower-arms}, and \textit{upper-/lower-legs}). It is split into 1,716 and 1,817 images for training and test. ATR~\cite{liang2015deep} is a challenging human parsing dataset, which has  7,700 single-person images with dense annotations over 17 categories (\eg, \textit{face, upper-clothes, left-/right-arms, left-/right-legs}, \etc). There are 6,000, 700 and 1,000 images for training, validation, and test, respectively.
PPSS~\cite{luo2013pedestrian} is a collection of 3,673 single-pedestrian images from 171 surveillance videos and provides pixel-wise annotations for \textit{hair, face, upper-/lower-clothes, arm}, and \textit{leg}.
It presents diverse real-word challenges, \eg, pose variations, illumination changes, and occlusions. There are 1,781 and 1,892 images for training and testing, respectively. {Fashion Clothing~\cite{Luo_2018_TGPnet}} has 4,371 images gathered from Colorful Fashion Parsing~\cite{liu2014fashion}, Fashionista~\cite{yamaguchi2012parsing}, and  Clothing Co-Parsing~\cite{yang2014clothing}. It has 17 clothing categories (\eg, \textit{hair, pants, shoes, upper-clothes}, \etc) and the data split follows 3,934 for training and 437 for test.

\noindent\textbf{Training:} ResNet101~\cite{he2016deep}, pre-trained on ImageNet~\cite{ImageNet}, is used to initialize our DeepLabV3\!~\cite{chen2018deeplabv3plus} backbone. The remaining layers are randomly initialized. We train our model on the five aforementioned datasets with their respective training samples, separately. Following the common practice~\cite{CE2P2019,gong2018instance,wang2019CNIF}, we randomly augment each training sample with a scaling factor in [0.5, 2.0], crop size of $473\!\times\!473$, and horizontal flip. %To enhance the model robustness of human appearance change, we also introduce random HSV (Hue, Saturation, Value), where the random hue shift is set from -18 degree to 18 degree, random saturation from 0.5 to 1.5, while random value shift from -30 to 30.
For optimization, we use the standard SGD solver, with a
momentum of 0.9 and weight\_decay of 0.0005.
To schedule the learning rate, we employ the
polynomial annealing procedure~\cite{chen2018deeplab,zhao2017pspnet}, where
the learning rate is multiplied by $(1\!-\!\frac{iter}{total\_iter})^{power}$ with \textit{power} as $0.9$. %and \textit{base\_lr}$=$$0.007$. The \textit{total\_iter} is $epochs\!\times\!batch\_size$, where \textit{batch\_size}$=$$40$~and~\textit{epochs}$=$$150$.
%We use multiple GPUs with Synchronized Cross-GPU Batch Normalization for the consumption of large batch size.

%We adopt different initial learning rates, batch sizes
%and training epochs by following the previous works. For
%Cityscapes, we choose initial learning rate of 0.01, batch
%size of 8 and crop size of 769 × 769 [5, 60]. For ADE20K,
%we choose initial learning rate of 0.02, batch size of 16 and
%crop size of 520 × 520 following [60, 53]. For LIP, we
%choose initial learning rate of 0.007, batch size of 40 and
%crop size of 473 × 473 following [27]. For PASCAL VOC
%2012 and PASCAL-Context, we choose initial learning rate
%of 0.01, batch size of 16 and crop size of 513 × 513 following [4, 55]. For COCO-Stuff, we choose initial learning rate
%of 0.01, batch size of 16 and crop size of 520×520. We train
%the models for 110 epochs on Cityscapes, 120 epochs on
%ADE20K, 150 on LIP, 80 epochs on PASCAL VOC 2012,
%PASCAL-Context and 100 epochs on COCO-Stuff.

\begin{table}[t]
\centering\small
\begin{threeparttable}
\setlength\tabcolsep{4pt}
\renewcommand\arraystretch{1.00}
\resizebox{0.42\textwidth}{!}{
\begin{tabular}{rIccc}    % {lccc}
%\hline\thickhline
%\rowcolor{mygray}
Method~~~~~ &~~pixAcc.~~ &Mean Acc. &~Mean IoU~\\
\hline\thickhline
SegNet~\cite{badrinarayanan2017segnet} &69.04 &24.00 &18.17\\
FCN-8s~\cite{long2015fully} &76.06 &36.75 &28.29\\
DeepLabV2~\cite{chen2018deeplab} &82.66 &51.64 &41.64\\
Attention~\cite{chen2016attention} &83.43 &54.39 &42.92\\
$^\dag$Attention+SSL~\cite{gong2017look}  &84.36 &54.94 &44.73\\
DeepLabV3+~\cite{chen2018deeplabv3plus} &84.09 &55.62 &44.80\\
ASN~\cite{luc2016semantic} &- &- &45.41\\
$^\dag$SSL~\cite{gong2017look} &- &- &46.19\\
MMAN~\cite{luo2018macro}  &85.24 &57.60 &46.93\\
$^\dag$SS-NAN~\cite{zhao2017self} &87.59 &56.03 &47.92\\
HSP-PRI~\cite{Kalayeh_2018_CVPR} &85.07 &60.54 &48.16\\
$^\dag$MuLA~\cite{nie2018mutual} &88.5 &60.5 &49.3\\
PSPNet~\cite{zhao2017pspnet} &86.23 &61.33 &50.56\\
CE2P~\cite{CE2P2019} &87.37 &63.20 &53.10\\
BraidNet~\cite{BraidNet} &87.60 &66.09 &54.42\\
CNIF~\cite{wang2019CNIF} &88.03 &68.80 &57.74 \\ \hline
\textbf{Ours} &\textbf{89.05}	  &\textbf{70.58} &\textbf{59.25} \\
%\hline
\end{tabular}
}
\end{threeparttable}
\vspace{-10pt}
\captionsetup{font=small}
\caption{\small\textbf{\!Comparison of pixel accuracy, mean accuracy and mIoU on LIP \texttt{val}}~\cite{gong2017look}. $^\dag$ indicates extra pose information used. %Please see the supplementary material for per-class performance. %(Higher values are better. The best score is marked in \textbf{bold}. These notes are the same for other tables.)
}
\label{tab:LIP1}
\vspace{-12pt}
\end{table}

\noindent\textbf{Testing:}  For each test sample, we set the long side of the image to 473 pixels and maintain the original aspect ratio.
As in~\cite{zhao2017pspnet,nie2018mutual}, we average the parsing results over five-scale image pyramids of different scales with flipping, \ie, the scaling factor is 0.5 to 1.5 (with intervals of 0.25).

\noindent\textbf{Reproducibility:} Our method is implemented on PyTorch and trained on four NVIDIA Tesla V100 GPUs (32GB memory per-card). All the experiments are performed on one NVIDIA TITAN Xp 12GB GPU. To provide full details of our approach, our code will be made publicly available.

\noindent\textbf{Evaluation:} For fair comparison, we follow the official evaluation protocols of each dataset. For LIP, following~\cite{zhao2017self}, we report pixel accuracy, mean accuracy and mean Intersection-over-Union (mIoU).
For PASCAL-Person-Part and PPSS, following~\cite{xia2016zoom,xia2017joint,luo2018macro}, the performance is evaluated in terms of mIoU. For ATR and Fashion Clothing, as in~\cite{Luo_2018_TGPnet,wang2019CNIF}, we report pixel accuracy, foreground accuracy, average precision, average recall, and average F1-score.

\vspace{-3pt}
\subsection{Quantitative and Qualitative Results}
\label{sec:qresults1}
\vspace{-2pt}

%We compare the proposed method with several strong baselines on the five aforementioned challenging datasets.
%
\noindent\textbf{LIP~\cite{gong2017look}:}$_{\!}$ LIP is a gold standard benchmark for human parsing. Table\!~\ref{tab:LIP1} reports the comparison results with 16 state-of-the-arts on LIP \texttt{val}. We first find that general semantic
segmentation methods~\cite{badrinarayanan2017segnet,long2015fully,chen2018deeplab,chen2018deeplabv3plus} %SegNet~\cite{badrinarayanan2017segnet}, FCN-8s~\cite{long2015fully}, DeepLabV2~\cite{chen2018deeplab}, DeepLabV3+~\cite{chen2018deeplabv3plus}
tend to perform worse than human parsers. This indicates the importance of reasoning human structures in this problem. In addition, though recent human parsers gain impressive results, our model still outperforms all the competitors by a large margin. For instance, in terms of pixAcc., mean Acc., and mean IoU, our parser dramatically surpasses the best performing method, CNIF~\cite{wang2019CNIF}, by 1.02\%, 1.78\% and 1.51\%, respectively.  We would also like to mention that our parser does not use additional pose~\cite{gong2017look,zhao2017self,nie2018mutual} or edge~\cite{CE2P2019} information. %Detailed analyses with per-class IoU performance can be found in the supplementary material.

%We further report per-class IoU on LIP validation set to
%verify the detailed effectiveness of our SS-NAN, presented
%in Table 2.  As observed from the reported results, SS-
%NAN significantly improves the performance of the labels
%like arms, legs, and shoes, which demonstrates its excel-
%lent ability to distinguish “left” v.s. “right”. Furthermore,
%the labels covering small regions such as sunglasses, gloves,
%socks, are predicted better with higher IoU. This improve-
%ment also verified the effectiveness of the proposed SS-
%NAN especially for small labels.
%%We can find the
%Our method achieves a huge boost in average IoU (4.64\% better than the second best method, CE2P~\cite{CE2P2019} and 8.4\% better
%than the third best, MuLA~\cite{nie2018mutual}). To verify its effectiveness in detail, we report per-class IoU in \autoref{tab:LIP2}. Our model improves the performance over almost all classes, especially for the ones typically associated with small regions (\eg, \textit{gloves, sunglasses, socks, shoes}), due to our top-down inference strategy. The results are also impressive for \textit{arms, legs}, and \textit{shoes}, demonstrating our model is able to distinguish between ``left'' and ``right'' with the help of composition relations.

\begin{table}%[H]
\centering\small
\begin{threeparttable}
\setlength\tabcolsep{2pt}
\renewcommand\arraystretch{1.00}
\resizebox{0.49\textwidth}{!}{
\begin{tabular}{rIcccccccIc}    % {lccc}
%\hline\thickhline
%\rowcolor{mygray}
Method~~~~~~~~~~&Head &Torso  &U-Arm &L-Arm &U-Leg &L-Leg &B.G. &Ave.\\
\hline\thickhline
HAZN~\cite{xia2016zoom} &80.79 &59.11 &43.05 &42.76 &38.99 &34.46 &93.59 &56.11\\
Attention~\cite{chen2016attention} &81.47 &59.06 &44.15 &42.50 &38.28 &35.62 &93.65 &56.39\\
LG-LSTM~\cite{liang2016semantic2} &82.72 &60.99 &45.40 &47.76 &42.33 &37.96 &88.63 &57.97\\
%Joint~\cite{xia2017joint} &80.21 &61.36 &47.53 &43.94 &41.77 &38.00 &93.64 &58.06\\
Attention+SSL~\cite{gong2017look} &83.26 &62.40 &47.80 &45.58 &42.32 &39.48 &94.68 &59.36\\
Attention+MMAN~\cite{luo2018macro} &82.58 &62.83 &48.49 &47.37 &42.80 &40.40 &94.92 &59.91\\
Graph LSTM~\cite{liang2016semantic} &82.69 &62.68 &46.88 &47.71 &45.66 &40.93 &94.59 &60.16\\
SS-NAN~\cite{zhao2017self}  &86.43 &67.28 &51.09 &48.07 &44.82 &42.15 &97.23 &62.44\\
Structure LSTM~\cite{liang2017interpretable} &82.89 &67.15 &51.42 &48.72 &51.72 &45.91 &97.18 &63.57\\
Joint~\cite{xia2017joint} &85.50 &67.87 &54.72 &54.30 &48.25 &44.76 &95.32 &64.39\\
DeepLabV2~\cite{chen2018deeplab} &- &- &- &- &- &- &- &64.94\\
MuLA~\cite{nie2018mutual} &84.6 &68.3 &57.5 &54.1 &49.6 &46.4 &95.6 &65.1\\
PCNet~\cite{Zhu2018ProgressiveCH} &86.81 &69.06 &55.35 &55.27 &50.21 &48.54 &96.07 &65.90\\
Holistic~\cite{li2017holistic} &86.00 &69.85 &56.63 &55.92 &51.46 &48.82 &95.73 &66.34\\
WSHP~\cite{fang2018weakly} &87.15&72.28&57.07&56.21&52.43&50.36&\textbf{97.72}&67.60\\
DeepLabV3+~\cite{chen2018deeplabv3plus} &87.02 &72.02 &60.37 &57.36 &53.54 &48.52 &96.07&67.84\\
SPGNet~\cite{Bowen2019} &87.67&71.41&61.69&60.35&52.62&48.80&95.98&68.36\\
PGN~\cite{gong2018instance} &\textbf{90.89} &75.12 &55.83 &64.61 &55.42 &41.57 &95.33 &68.40\\
CNIF~\cite{wang2019CNIF} & 88.02 &72.91 &{64.31} &63.52 &{55.61} &{54.96} &96.02 &70.76 \\
%Graphonomy~\cite{Gong_2019_CVPR}&88.04 &74.99 &61.49 &60.72 &54.21 &50.06&95.87 &71.14\\
%DPC~\cite{chen2018searching}&88.81 &74.54 &63.85 &63.73 &57.24 &54.55 &96.66 &71.34\\
\hline
\textbf{Ours} &89.73	&\textbf{75.22}	&\textbf{66.87}	&\textbf{66.21}	&\textbf{58.69}	&\textbf{58.17}	&{96.94}	&\textbf{73.12}\\
%\hline
\end{tabular}
}
\end{threeparttable}
\vspace{-11pt}
\captionsetup{font=small}
\caption{\small \textbf{Per-class comparison of mIoU on PASCAL-Person-Part \texttt{test}}~\cite{xia2017joint}.}
\label{tab:PASCAL-Person-Part}
\vspace{-12pt}
\end{table}

\begin{table}[t]
\centering\small
\begin{threeparttable}
\setlength\tabcolsep{2pt}
\renewcommand\arraystretch{1.00}
\resizebox{0.49\textwidth}{!}{
\begin{tabular}{rIccccIc}    % {lccc}
%\rowcolor{mygray}
Method~~~~  &~~~pixAcc.~~~  &F.G. Acc. &~~~Prec.~~~ &~~~Recall~~~ &~~~F-1~~~\\
\hline\thickhline
Yamaguchi~\cite{yamaguchi2012parsing} &84.38 &55.59 &37.54 &51.05 &41.80 \\
Paperdoll~\cite{yamaguchi2013paper} &88.96 &62.18 &52.75 &49.43 &44.76 \\
M-CNN~\cite{liu2015matching} &89.57 &73.98 &64.56 &65.17 &62.81 \\
ATR~\cite{liang2015deep} &91.11 &71.04 &71.69 &60.25 &64.38 \\
DeepLabV2~\cite{chen2018deeplab} &94.42 &82.93 &78.48 &69.24 &73.53 \\
PSPNet~\cite{zhao2017pspnet} &95.20 &80.23 &79.66 &73.79 &75.84 \\
Attention~\cite{chen2016attention} &95.41 &85.71 &81.30 &73.55 &77.23 \\
DeepLabV3+~\cite{chen2018deeplabv3plus} &95.96 &83.04 &80.41 &78.79 &79.49 \\
Co-CNN~\cite{liang2015human} &96.02 &83.57 &{84.95} &77.66 &80.14 \\
LG-LSTM~\cite{liang2016semantic2} &96.18 &84.79 &84.64 &79.43 &80.97\\
TGPNet~\cite{Luo_2018_TGPnet} &{96.45} &{87.91} &83.36 &80.22 &81.76 \\
%Graph LSTM~\cite{liang2016semantic}  &97.60 &91.42 &84.74 &83.28 &83.76\\
CNIF~\cite{wang2019CNIF} &96.26 &{87.91} &84.62 &{86.41} &{85.51}\\
%Structure LSTM~\cite{liang2017interpretable}&97.71 &91.76 &89.37 &86.84 &87.88\\
%Graphonomy~\cite{Gong_2019_CVPR}&98.32 &92.94 &88.14 &89.97  &90.89\\
 \hline
\textbf{Ours} &\textbf{96.84}	&\textbf{89.23}	&\textbf{86.17}	&\textbf{88.35}	&\textbf{87.25} \\
\end{tabular}
}
\end{threeparttable}
\vspace{-11pt}
\captionsetup{font=small}
\caption{\small\textbf{\!Comparison of accuracy, foreground accuracy, average precision, recall and F1-score on ATR \texttt{test}}\!~\cite{liang2015deep}. %Please see the supplementary material for per-class performance.
}
\label{tab:ATR1}
\vspace{-12pt}
\end{table}

\noindent\textbf{PASCAL-Person-Part~\cite{xia2017joint}:} In Table\!~\ref{tab:PASCAL-Person-Part}, we compare our method against 18 recent methods on PASCAL-Person-Part \texttt{test} using IoU score.  From the results, we can again see that our approach achieves better performance compared to all other methods; specially,  73.12\% \textit{vs} 70.76\% of CNIF~\cite{wang2019CNIF} and 68.40\% of PGN~\cite{gong2018instance}, in
terms of \textit{mIoU}. Such a performance gain is particularly impressive considering that improvement on this dataset is very challenging.

%our model outperforms previous methods across the vast majority of classes and on average.
%
\noindent\textbf{ATR~\cite{liang2015deep}:} Table\!~\ref{tab:ATR1} presents comparisons
with 14 previous methods on ATR \texttt{test}. Our approach sets new state-of-the-arts for all five metrics,
outperforming all  other methods by a large margin. For example, our parser provides a considerable performance gain in F-1 score, \ie, 1.74\% and 5.49\% higher than the current top-two performing methods, CNIF~\cite{wang2019CNIF} and TGPNet~\cite{Luo_2018_TGPnet}, respectively.

\noindent\textbf{Fashion Clothing~\cite{Luo_2018_TGPnet}:} The quantitative comparison results with six competitors on Fashion Clothing \texttt{test} are summarized in Table~\ref{tab:Clothing}. Our model yields an F-1 score of~60.19\%, while those for Attention~\cite{chen2016attention}, TGPNet~\cite{Luo_2018_TGPnet}, and CNIF~\cite{wang2019CNIF} are 48.68\%, 51.92\%, and 58.12\%, respectively. This again demonstrates our superior performance.

\begin{table}[t]
\centering\small
\begin{threeparttable}
\setlength\tabcolsep{2pt}
\renewcommand\arraystretch{1.00}
\resizebox{0.49\textwidth}{!}{
\begin{tabular}{rIccccIc}    % {lccc}
%\rowcolor{mygray}
Method~~~~  &~~pixAcc.~~  &F.G. Acc. &~~~Prec.~~~ &~~~Recall~~~ &~~~F-1~~~\\
\hline\thickhline
Yamaguchi~\cite{yamaguchi2012parsing} &81.32 &32.24 &23.74 &23.68 &22.67 \\
Paperdoll~\cite{yamaguchi2013paper} &87.17 &50.59 &45.80 &34.20 &35.13 \\
DeepLabV2~\cite{chen2018deeplab} &87.68 &56.08 &35.35 &39.00 &37.09 \\
Attention~\cite{chen2016attention} &90.58 &64.47 &47.11 &50.35 &48.68 \\
TGPNet~\cite{Luo_2018_TGPnet} &91.25 &66.37 &50.71 &53.18 &51.92 \\
CNIF~\cite{wang2019CNIF} &92.20 &{68.59} &{56.84} &{59.47} &58.12 \\\hline
\textbf{Ours} &\textbf{93.12}	&\textbf{70.57}	&\textbf{58.73}	&\textbf{61.72}	&\textbf{60.19} \\
%\hline
\end{tabular}
}
\end{threeparttable}
\vspace{-10pt}
\captionsetup{font=small}
\caption{\small \textbf{Comparison of pixel accuracy, foreground pixel accuracy, average precision, average recall and average f1-score on Fashion Clothing \texttt{test}}~\cite{Luo_2018_TGPnet}.}
\label{tab:Clothing}
\vspace{-8pt}
\end{table}

\begin{table}[t]
\centering\small
\begin{threeparttable}
\setlength\tabcolsep{2pt}
\renewcommand\arraystretch{1.00}
\resizebox{0.49\textwidth}{!}{
\begin{tabular}{rIcccccccIc}    % {lccc}
%\rowcolor{mygray}
Method~~ &Head &Face  &U-Cloth &Arms &L-Cloth &Legs &~B.G.~ &~Ave.~\\
\hline\thickhline
DL~\cite{luo2013pedestrian} &22.0 &29.1 &57.3 &10.6 &46.1 &12.9 &68.6 &35.2\\
DDN~\cite{luo2013pedestrian} &35.5 &44.1 &68.4 &17.0 &61.7 &23.8 &80.0 &47.2\\
ASN~\cite{luc2016semantic} &51.7 &51.0 &65.9 &29.5 &52.8 &20.3 &83.8 &50.7\\
MMAN~\cite{luo2018macro} &53.1 &50.2 &69.0 &29.4 &55.9 &21.4 &85.7 &52.1\\
LCPC~\cite{Dang_BMVC}&55.6 &46.6 &71.9 &30.9 &58.8 &24.6 &86.2 &53.5\\
CNIF~\cite{wang2019CNIF} &{67.6} &{60.8} &{80.8} &{46.8} &{69.5} &{28.7} &{90.6} &{60.5} \\ \hline
\textbf{Ours} &\textbf{68.8}	&\textbf{63.2}	&\textbf{81.7}	&\textbf{49.3}	&\textbf{70.8}	&\textbf{32.0}	&\textbf{91.4}	&\textbf{65.3}\\
\end{tabular}
}
\end{threeparttable}
\vspace{-8pt}
\captionsetup{font=small}
\caption{\small \textbf{Comparison of mIoU on PPSS \texttt{test}}~\cite{luo2013pedestrian}.}
\label{tab:PPSS}
\vspace{-16pt}
\end{table}

 %representative

\noindent\textbf{PPSS~\cite{luo2013pedestrian}:} Table~\!\ref{tab:PPSS} compares our method against six famous methods on PPSS \texttt{test} set. The evaluation results demonstrate
that our human parser achieves 65.3\% mIoU, with substantial gains over the second best, CNIF~\cite{wang2019CNIF}, and third best, LCPC~\cite{Dang_BMVC}, of 4.8\% and 11.8\%, respectively.

%%%%%%%%%%%%%%%%%%% Figure 2%%%%%%%%%%%%%%%%%%%%%%
\begin{figure*}[t]
%%tr = 0.006, ts = 0.008
  \centering
      \includegraphics[width=1 \linewidth]{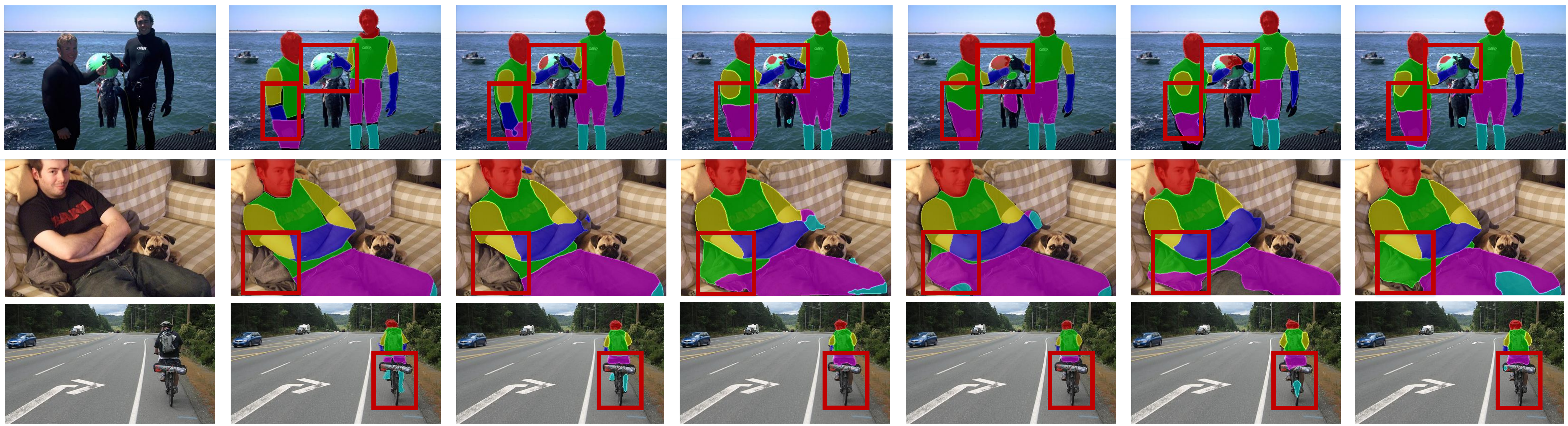}
%    \put(-479,182){\footnotesize(a) Image}
%    \put(-420,182){\footnotesize(b) Ground-truth}
%    \put(-337,182){\footnotesize(c) Ours}
%    \put(-282,182){\footnotesize(d) DeepLabV3+~\cite{chen2018deeplabv3plus}}
%	\put(-200,182){\footnotesize(e) PGN~\cite{gong2018instance}}
%    \put(-135,182){\footnotesize(f) SS-NAN~\cite{zhao2017self}}
%	\put(-58 ,182){\footnotesize(g) CNIF~\cite{wang2019CNIF}}
	\put(-479,-7 ){\footnotesize(a) Image}
	\put(-420,-7 ){\footnotesize(b) Ground-truth}
    \put(-337,-7 ){\footnotesize(c) Ours}
    \put(-282,-7 ){\footnotesize(d) DeepLabV3+~\cite{chen2018deeplabv3plus}}
	\put(-200,-7 ){\footnotesize(e) PGN~\cite{gong2018instance}}
    \put(-135,-7 ){\footnotesize(f) SS-NAN~\cite{zhao2017self}}
	\put(-58 ,-7 ){\footnotesize(g) CNIF~\cite{wang2019CNIF}}
\vspace{-6pt}
\captionsetup{font=small}
\caption{\small\textbf{Visual comparison on PASCAL-Person-Part \texttt{test}.} Our model (c) generates more accurate predictions, compared to other famous methods~\cite{chen2018deeplabv3plus,gong2018instance,zhao2017self,wang2019CNIF} (d-g). The improved labeled results by our parser are denoted in red boxes. Best viewed in color.%See \S\ref{sec:qresults2} for details.
}
\label{fig:visual}
\vspace{-15pt}
\end{figure*}
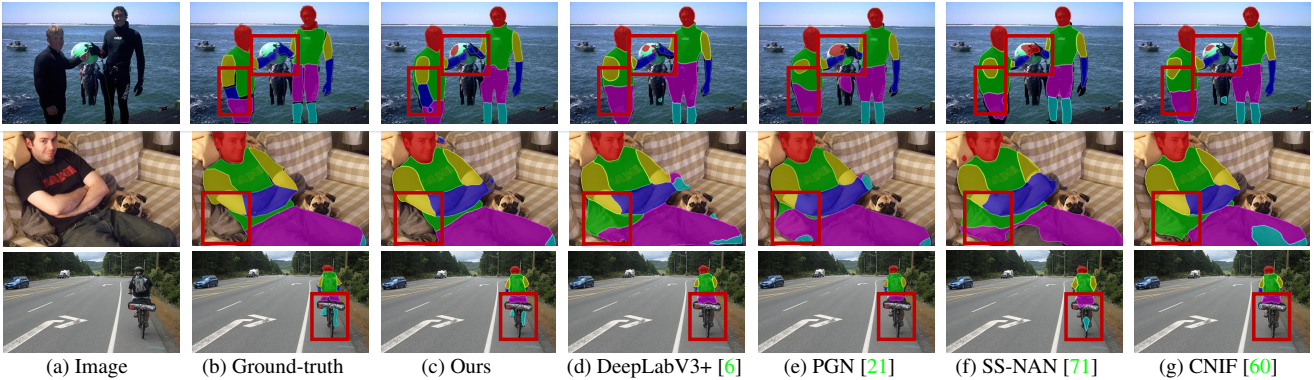

\noindent\textbf{Runtime$_{\!}$ comparison:}$_{\!}$ As$_{\!}$ our$_{\!}$ parser$_{\!}$ does$_{\!}$ not$_{\!}$ require$_{\!}$ extra pre-/post-processing steps (\eg, human pose used in~\cite{xia2017joint}, over-segmentation in~\cite{liang2016semantic,liang2017interpretable}, and CRF in~\cite{xia2017joint}), it achieves a high speed of 12fps (on PASCAL-Person-Part), faster than most of the counterparts, such as Joint~\cite{xia2017joint} (0.1fps), Attention+SSL~\cite{gong2017look} (2.0fps), MMAN~\cite{luo2018macro} (3.5fps), SS-NAN~\cite{zhao2017self} (2.0fps), and LG-LSTM~\cite{liang2016semantic2} (3.0fps).

\noindent\textbf{Qualitative results:} Some qualitative comparison results on PASCAL-Person-Part \texttt{test} are depicted in Fig.~\!\ref{fig:visual}. We can see that our approach outputs more precise parsing results than other
competitors~\cite{chen2018deeplabv3plus,gong2018instance,zhao2017self,wang2019CNIF},
despite the existence of rare pose (2$_{nd}$ row) and occlusion (3$_{rd}$ row). In addition, with its better understanding of human structures, our parser gets more robust results and eliminates the interference from the background (1$_{st}$ row). The last row gives a challenging case, where our parser still correctly recognizes the confusing parts of the person in the middle.

Overall, our human parser attains strong performance across all the five datasets. We believe this is due to our typed relation modeling and iterative algorithm, which enable more trackable part features  and better approximations. %In \S\ref{sec:ablation}, we will ablatively study the contribution of each essential components of our model, enabling further detailed and in-depth analyses.

\vspace{-2pt}
\subsection{Diagnostic Experiments}
\label{sec:ablation}
\vspace{-1pt}
%The incorporation of all components leads to a significant improvement on the Buffy data se

To demonstrate how each component in our parser contributes to the performance, a series of ablation experiments are conducted on PASCAL-Person-Part \texttt{test}. %The training and evaluation followed the same protocol as in \S\ref{sec:dataset}. %The experimental results are summarized in Table~\ref{tab:ablation}.

\noindent\textbf{Type-specific relation modeling:} We first investigate the necessity of comprehensively exploring different relations, and discuss the effective of our type-specific relation modeling strategy. Concretely, we studied six variant models, as$_{\!}$ listed$_{\!}$ in$_{\!}$ Table~\!\ref{tab:ablation}:$_{\!}$ \textbf{(1)}$_{\!}$ `Baseline'$_{\!}$ denotes$_{\!}$ the$_{\!}$ approach$_{\!}$ only$_{\!}$ using$_{\!}$ the$_{\!}$ initial$_{\!}$ node$_{\!}$ embeddings$_{\!}$ $\{\textbf{\textit{h}}^{(0)\!}_v\}_{v\in\mathcal{V}\!}$ without$_{\!}$ any relation information; \textbf{(2)} `Type-agnostic' shows the performance when modeling different human part relations in a type-agnostic manner: $\!\textbf{\textit{h}}_{u,v\!}\!\!=_{\!}\!\!R([\textbf{\textit{h}}_{u}, \textbf{\textit{h}}_v])$; \textbf{(3)} `Type-specific \textit{w/o} $F^r$' gives the performance without the relation-adaption operation $F^{r\!}$ in Eq.~\!\ref{eq:edge-specific}:$_{\!}$ $\textbf{\textit{h}}_{u,v\!}\!\!=_{\!}\!\!R^r([\textbf{\textit{h}}_u, \textbf{\textit{h}}_v])$;
\textbf{(4-6)} `Decomposition relation', `Composition relation' and `Dependency relation' are three variants that only consider the corresponding single one of the three kinds of relation categories, using our type-specific relation modeling strategy (Eq.~\!\ref{eq:edge-specific}). Four main conclusions can be drawn: \textbf{(1)} Structural information are essential for human parsing, as all the structured models outperforms `Baseline'. \textbf{(2)} Typed relation modeling leads to more effective human structure learning, as `Type-specific \textit{w/o} $F^r$' improves `Type-agnostic' by 1.28\%. \textbf{(3)} Exploring different kinds of relations are meaningful, as the variants using individual relation types outperform `Baseline' and our full model considering all the three kinds of relations achieves the best performance. \textbf{(4)} Encoding relation-specific constrains helps with relation pattern learning as our full model is better than the one without relation-adaption, `Type-specific \textit{w/o} $F^r$'. %More ablation studies can be found in the supplementary material.

\begin{table}[t]
\centering\small
\begin{threeparttable}
\setlength\tabcolsep{4pt}
\renewcommand\arraystretch{1.00}
\resizebox{0.48\textwidth}{!}{
\begin{tabular}{cIcIccIc}    % {lccc}
Component&	Module &mIoU &$\triangle$ &time (ms)\\
\hline\thickhline
			Reference&\textbf{Full model} (2 iterations)  & 73.12   &- &81\\
            \hline
			\multirow{6}{*}{\tabincell{c}{Relation\\modeling}}
            &Baseline                           &68.84 & -4.28 &46\\
			&Type-agnostic                      &70.37 & -2.75 &55\\
			&Type-specific \textit{w/o} $F^r$   &71.65 & -1.47 &55\\
            &Decomposition relation              &71.38 & -1.74 &50\\
            &Composition relation           &69.35 & -3.77 &49\\
            &Dependency relation             &69.43 & -3.69 &52\\
\hline
			\multirow{6}{*}{\tabincell{c}{Iterative\\Inference $T$} }
            &0 iteration~ &68.84 & -4.28 &46\\
			&1 iterations &72.17 & -0.95 &59\\
			&3 iterations &73.19 & +0.07 &93\\
            &4 iterations &73.22 & +0.10 &105\\
            &5 iterations &73.23 & +0.11 &116\\
\end{tabular}
}
\end{threeparttable}
\vspace{-8pt}
\captionsetup{font=small}
\caption{\small \textbf{Ablation study (\S\ref{sec:ablation}) on PASCAL-Person-Part \texttt{test}.}}
\label{tab:ablation}
\vspace{-14pt}
\end{table}

\noindent\textbf{Iterative inference:} Table~\ref{tab:ablation} shows the performance of our parser with regard to the iteration step $t$ as denoted in
Eq.~\!\ref{eq:message} and Eq.~\!\ref{eq:update}. Note that, when $t\!=\!0$, only the initial node feature is used.  It can be observed that setting $T\!=\!2$ or $T\!=\!3$ provided a consistent boost in accuracy of 4$\sim$5\%, on average, compared to $T\!=\!0$; however, increasing $T$ beyond 3 gave marginal returns in performance (around 0.1\%). Accordingly, we choose $T\!=\!2$ for a better trade-off between accuracy and computation time.

\vspace{-3pt}
\section{Conclusion}
\vspace{-1pt}
In the human semantic parsing task, structure modeling is an essential, albeit
inherently difficult, avenue to explore. This work proposed a hierarchical human parser that addresses this issue in two aspects. First, three distinct relation networks are designed to precisely describe the compositional/decompositional relations between constituent and entire parts and help with the dependency learning over kinetically connected parts. Second, to address the inference over the loopy human structure, our parser relies on a convolutional, message passing based approximation algorithm, which enjoys the advantages of iterative optimization and spatial information preservation. The above designs enable strong performance across five widely adopted benchmark datasets, at times outperforming all other competitors.  %The above designs enable strong performance across five widely adopted human parsing benchmark datasets, at times outperforming all other competitors.
% For notational clarity

%{\small\noindent\textbf{Acknowledgements} This work was partially sponsored by Zhejiang Lab's Open Fund (No.~2019KD0AB04), Zhejiang Lab's International Talent$_{\!}$ Fund$_{\!}$ for$_{\!}$ Young$_{\!}$ Professionals, and$_{\!}$ CCF-Tencent$_{\!}$ Open$_{\!}$ Fund.$_{\!}$$_{\!}$  This work was also partially supported the National Key R\&D Program of China (Grant Nos. 2018AAA0102800 and 2018AAA0102802) and National Natural Science Foundation of China (Grant No. 61632018).

{\small
\bibliographystyle{ieee_fullname}
\bibliography{egbib}
}

\end{document}